%% file: main.tex
\pdfoutput=1
\documentclass[acmsmall, nonacm]{acmart}

\AtBeginDocument{%
  \providecommand\BibTeX{{%
    \normalfont B\kern-0.5em{\scshape i\kern-0.25em b}\kern-0.8em\TeX}}}

\usepackage{subfig}
\usepackage{comment}
\usepackage{textcomp}
\usepackage{hyperref}
\usepackage{epstopdf}
\usepackage{url}
\usepackage{graphicx}
\usepackage{bbding}
\usepackage{multirow}
\newcommand{\ie}{i.e.}
\newcommand{\eg}{e.g.}

\newcommand{\et}{et al.}
\newcommand{\tabincell}[2]{\begin{tabular}{@{}#1@{}}#2\end{tabular}}
\DeclareMathOperator*{\argmin}{arg\,min}

\usepackage{lipsum}
\newcommand\blfootnote[1]{%
  \begingroup
  \renewcommand\thefootnote{}\footnote{#1}%
  \addtocounter{footnote}{-1}%
  \endgroup
}

\begin{document}

\title{From Server-Based to Client-Based Machine Learning: A Comprehensive Survey}

\author{Renjie~Gu}
\email{grj165@sjtu.edu.cn}
\orcid{0003-2895-444X}
\author{Chaoyue~Niu}
\email{rvince@sjtu.edu.cn}
\author{Fan~Wu}
\email{fwu@cs.sjtu.edu.cn}
\authornote{Fan Wu is the corresponding author.}
\author{Guihai~Chen}
\email{gchen@cs.sjtu.edu.cn}
\affiliation{%
  \institution{Shanghai Jiao Tong University}
  \streetaddress{800 Dongchuan Rd.}
  \city{Shanghai}
  \country{China}
  \postcode{200240}
}
\author{Chun~Hu}
\email{shiji.hc@alibaba-inc.com}
\author{Chengfei~Lyu}
\email{chengfei.lcf@alibaba-inc.com}
\author{Zhihua~Wu}
\email{zhihua.wzh@alibaba-inc.com}
\affiliation{%
  \institution{Alibaba Group}
  \streetaddress{969 Wenyi Rd. (W)}
  \city{Hangzhou}
  \country{China}
  \postcode{311121}
}

\renewcommand{\shortauthors}{Gu et al.}

\begin{abstract}
In recent years, mobile devices have gained increasing development with stronger computation capability and larger storage space. Some of the computation-intensive machine learning tasks can now be run on mobile devices. To exploit the resources available on mobile devices and preserve personal privacy, the concept of client-based machine learning has been proposed. It leverages the users' local hardware and local data to solve machine learning sub-problems on mobile devices and only uploads computation results rather than the original data for the optimization of the global model. Such an architecture can not only relieve computation and storage burdens on servers but also can protect the users' sensitive information. Another benefit is the bandwidth reduction because various kinds of local data can be involved in the training process without being uploaded. In this paper, we provided a literature review on the progressive development of machine learning from server-based to client-based. We revisited a number of widely-used server-based and client-based machine learning methods and applications. We also extensively discussed the challenges and future directions in this area. We believe that this survey will give a clear overview of client-based machine learning and provide guidelines on applying client-based machine learning to practice.
\end{abstract}



\blfootnote{Comments: Accepted to ACM CSUR 2021, Volume 54, Issue 1, Pages 6:1 - 6:36. Webpage: \url{https://doi.org/10.1145/3424660}.}
\blfootnote{This work was supported in part by National Key R\&D Program of China No. 2019YFB2102200, in part by China NSF grant No. 61972252, 61972254, 61672348, and 61672353, in part by Joint Scientific Research Foundation of the State Education Ministry No. 6141A02033702, and in part by Alibaba Group through Alibaba Innovation Research Program. The opinions, findings, conclusions, and recommendations expressed in this paper are those of the authors and do not necessarily reflect the views of the funding agencies or the government. The authors also want to sincerely thank Dr. Shuo Yang for offering valuable suggestions on polishing the initial version of this work.}

\maketitle
\input{1_introduction}
\input{2_centralized_machine_learning}
\input{3_distributed_training}
\input{4_client_inference}
\input{5_client_training}
\input{6_future_directions}
\input{7_conclusion}




\bibliographystyle{ACM-Reference-Format}
\bibliography{reference}

\end{document}

%% file: 1_introduction.tex
\section{Introduction} \label{introduction}

Machine learning, especially deep learning, has become a hot topic, attracting tremendous attention from both academia and industry. The core idea of machine learning is to use large amounts of data to train a model that can generalize well to unseen test samples. However, with the increase of data volume and the enhancement of model capacity, it is infeasible for a single server to accomplish complex learning tasks in a centralized way. To address this problem, the concept of server-based distributed machine learning was proposed in~\cite{provost1996scaling}, where multiple servers, connected through shared data buses or a fast local area network, exchange essential information (\eg, training losses and gradients) to collaboratively train a model. Although this framework is highly scalable and has been widely deployed in practice, it may not always be cost effective and efficient to build a high-performance server cluster. In addition to cost, security and privacy are major concerns when machine learning involves sensitive user data, such as typed texts in natural language processing, user profiles in personalized recommendations, and health records in medical diagnosis. Specifically, the servers in both centralized and distributed machine learning frameworks require direct accesses to training data and thus need to collect and store user data, which inevitably suffers outsider and insider attacks~\cite{popa2011cryptdb, popa2012cryptdb, popa2014building}. For example, a malicious hacker may invade the datacenter, compromise part of servers, and leak private databases. Further, if the servers are untrusted, they may share user data with other unauthorized entities or even trade for profits. In a nutshell, how to reduce the server cluster's operation cost, how the trusted servers can securely maintain user data, and how to defend untrusted servers are bottlenecks of the server-based machine learning.

Meanwhile, with rapid proliferation and development of mobile devices, the idea of doing machine learning tasks on mobile devices has also emerged. For example, applications, such as face recognition and speech recognition, are all based on machine learning and are common among mobile phones. To support these applications, a full-sized machine learning model is first trained on servers using large amounts of data, and then it is tailored and delivered to mobile devices to do inference and make predictions locally. This framework brings all the burdens to the central servers, wasting the resources of mobile devices, whose processors, memory space, and disk space are now powerful and abundant enough to support various kinds of computation tasks. In addition, many off-the-shelf machine learning frameworks (\eg, the TensorFlow Lite module in TensorFlow~\cite{tensorflow2015-whitepaper}) are now available. Developers can now readily adopt these end-to-end tools to build machine learning models for their mobile applications. The above evidences have shown that it is feasible to deploy distributed training tasks on mobile devices, which is also called client-based training.

Client-based training has advantages in cost reduction and privacy preservation. In particular, machine learning problems are distributed to mobile devices and solved locally so that high-performance servers and user data transmission/maintenance are no longer required. The idea of client-based training can be traced back to 2015, when Shokri~\et~\cite{shokri2015privacy} proposed distributed deep learning without sharing datasets among multiple parties. Later that year, Google researchers designed federated optimization~\cite{konevcny2015federated}, aiming to improve communication efficiency during learning with decentralized datasets. The idea was also referred to as Federated Learning and further developed in the following years~\cite{mcmahan2016communication, konevcny2016federated, konevcny2016federated2, konecny2017stochastic}. These works can be generally viewed as a specific type of client-based training which mainly focuses on how to make use of data without uploading them to the server, so that the privacy of users can be well preserved. If we 
enable on-device training and only upload the computation results, the leakage of sensitive information can be relieved. The reason is that attacks against the computation results without accessing the raw data are much harder. Moreover, since raw data is processed locally, client-based training is now able to make use of the data that is too much to be uploaded (so that centralized machine learning doesn't take them into consideration), which gives us great opportunity to improve the model performance.

Compared to existing surveys in this area, this survey focuses on the evolution process of machine learning. From server-based machine learning to client-based machine learning, the application scenarios and the research problems have greatly changed. We summarize the changes and further investigate the underlying motivations. We also review the research focuses at different stages of machine learning development. In particular, considering that a lot of new features and new demands have emerged in client-based machine learning, we analyze the applicability of existing server-based algorithms to client-based machine learning. We finally point out some future directions of client-based machine learning. In a nutshell, this survey not only is a review of the development of machine learning, but also can work as a good reference for designing new client-based learning algorithms on the basis of conventional server-based methods.

The rest of this survey is organized as follows. In Section \ref{sec:centralized_machine_learning} and Section \ref{sec:server_distributed_training}, we introduce the general machine learning process and the methodology of server-based distributed training. In Section \ref{sec:client_inference} and Section \ref{sec:client_training}, we explain the motivations of client-based machine learning (including inference and training), list the challenges, and discuss current advances of client-based training using federated learning and split learning as examples. In Section \ref{sec:future_directions}, we discuss the open problems and future directions of client-based training. Finally, we conclude the survey in Section \ref{sec:conclusion}.

%% file: 2_centralized_machine_learning.tex
\section{Centralized Machine Learning} \label{sec:centralized_machine_learning}

Machine learning~\cite{bishop2006pattern} is a study of mathematical models that can automatically learn and make predictions based on a set of observed data. The concept of centralized machine learning means that operations and executions of the model are all done on a central machine. Centralized machine learning has been widely used to extract insights behind huge amounts of data.

In this section, we briefly review concepts and techniques of centralized machine learning. To better understand the methodology of machine learning, we first introduce the paradigms of machine learning methods in Section \ref{subsec:machine_learning_paradigms}. Next, several concepts that are commonly used in machine learning will be illustrated in Section \ref{subsec:machine_learning_concepts}. Then, the task of machine learning will be defined in Section~\ref{subsec:machine_learning_task_definition}. A general process of machine learning will be presented in Section \ref{subsec:machine_learning_process}. After that, several machine learning optimizers will be introduced in Section~\ref{subsec:gdbased_optimizers}. Finally, the applicability of these optimizers to client-based training will be discussed in Section~\ref{subsec:applicability_of_gdbased_optimizers}.

\subsection{Machine Learning Paradigm} \label{subsec:machine_learning_paradigms}
\begin{figure}[t]
	\centering
	\subfloat[Machine Learning]{\label{fig:machine_learning_taxonomy}
	\begin{minipage}{.55\linewidth}
	\centering
	\includegraphics[width=\textwidth]{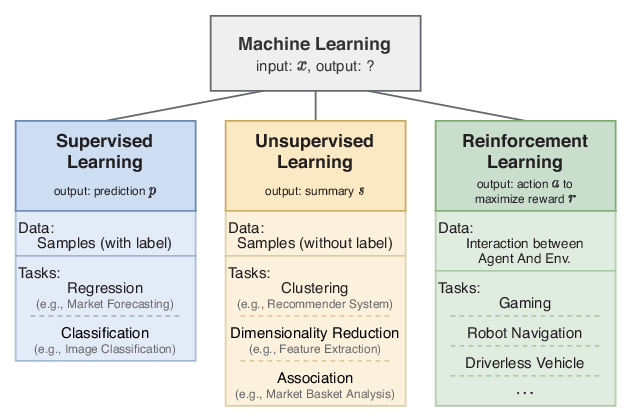}
	\end{minipage}
	}
	\subfloat[Supervised Learning]{\label{fig:supervised_learning_taxonomy}
	\begin{minipage}{.45\linewidth}
	\centering
	\includegraphics[width=\textwidth]{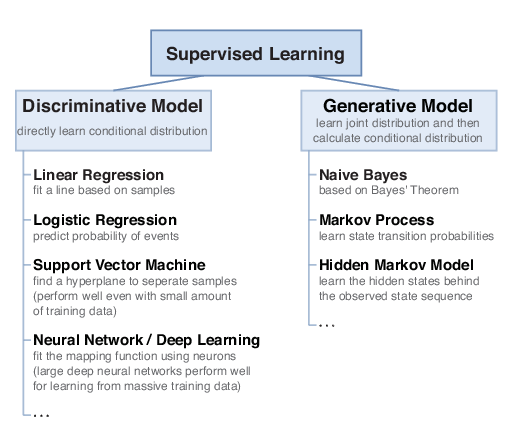}
	\end{minipage}
	}
	\caption{Taxonomy of machine learning and supervised learning.}\label{fig:taxonomy}
\end{figure}

If we classify machine learning algorithms based on the kind of input data samples and the kind of output, we have the following three basic machine learning paradigms as shown in Fig.~\ref{fig:machine_learning_taxonomy}.

\paragraph{\textbf{Supervised Learning}} In supervised learning, every data sample is made up of several input features and a label. The learning process is to approximate a mapping function from the features to the label. After that, given new input features, the label for the data can be predicted using the mapping function. This scheme is the most popular machine learning scheme which has been used in a variety of tasks. An example of supervised learning is the classification task, which is to classify an object based on its features, such as classifying fruit according to its color, shape, and weight. If the supervised learning task is to predict a continuous variable such as market pricing, then this is a regression task. We can further classify supervised learning based on the model type, as shown in Fig.~\ref{fig:supervised_learning_taxonomy}. We mainly focus on supervised learning~(especially discriminative models) in this survey.

\paragraph{\textbf{Unsupervised Learning}} In contrast to supervised learning, unsupervised learning is where we only have input features but no corresponding labels. Thus, the goal for unsupervised learning is to learn the distribution of the data and show how the data points are different from each other. A typical example of unsupervised learning is the clustering problem, which is to discover the groupings of the data, such as grouping users by their behaviors.

\paragraph{\textbf{Reinforcement Learning}} Reinforcement learning is quite different from supervised learning and unsupervised learning. Labelled input/output pairs and explicit correction on sub-optimal options are not needed for training an agent using reinforcement learning. Instead, the agent tries to find a trade-off between exploration and exploitation through its interaction with the environment. For good choices or actions, the agent gains rewards from the interpreter. Otherwise, it is punished. Reinforcement learning is widely used in research about robots or computer gaming agents. 

\paragraph{\textbf{Others}} In addition to the three basic paradigms, there also exists some other paradigms that can be viewed as a combination of the basic ones~(\eg, semi-supervised learning). Usually, the choice of machine learning paradigms depends on the kind of the problem going to be solved.

\subsection{Machine Learning Concepts} \label{subsec:machine_learning_concepts}
To better understand machine learning, in this section, we introduce and illustrate several key concepts that are commonly used in machine learning~(especially supervised learning).

\paragraph{\textbf{Stage and Dataset}} The whole process of machine learning is mainly made up of three kinds of stages, which are training, validation, and test. The dataset $D = \{d_1, d_2, \dots, d_n\}$ is the set of all $n$ data samples used for the machine learning process. A data sample $d_i = (x_i, y_i)$ contains a feature vector $x_i$ and a corresponding label $y_i$. Usually $D$ can be divided into three disjoint sub-sets which are the training set $D_{train}$, the validation set $D_{vld}$, and the test set $D_{test}$, corresponding to the three stages. $D_{train}$ accounts for the largest proportion of $D$. In the training stage, loss is calculated on $D_{train}$ and then optimization to the model is done. The validation stage aims to prevent overfitting. Since samples in $D_{vld}$ are not used in any training iterations, the error calculated on them will increase significantly if the model overfits $D_{train}$. Moreover, $D_{vld}$ is also helpful for tuning hyperparameters between training epochs. The test stage is to evaluate the performance of the final model on $D_{test}$. Thus, a typical machine leaning process is as follows: (1) Run training and validation iteratively until the model converges; and (2) Run test to evaluate the model performance.

\paragraph{\textbf{Feature}} In machine learning, a feature vector $x$ is a $k$-dimensional vector containing numerical features which are observable or measurable properties of instances, objects, or phenomena. We input feature vectors to the machine so that it knows how instances are different from each other.

\paragraph{\textbf{Label}} A label $y$ is the identity of the instance. Unlike features that describe the instances to the machine, labels tell the machine what the instances really are.

\paragraph{\textbf{Model}} The machine learning model $f(w;\cdot)$ includes a structure $f$ and a parameter vector $w$. The model $f(w;\cdot)$ is the core of machine learning. It serves as the mapping function that maps the input feature vector $x$ to the output label $f(w;x)$. For a good model, its output label $f(w;x)$ is close to the true label $y$ of the feature vector $x$. Usually, the structure and the parameters of the model are stored as multiple vectors or matrices. The mapping process is done through matrix multiplication. Various kinds of models such as support vector machines and artificial neural networks have been designed for different machine learning tasks.

\paragraph{\textbf{Loss Function}} The loss function $L(y,f(w;x))$ (\eg, mean square error, hinge loss, and softmax loss) describes how far the predicted label $f(w;x)$ deviates from the true label $y$. Since a pair $(x,y)$ is also known as a data sample $d$, we can also denote the loss function by $L(w;d)$ for convenience. The design of the loss function $L$ reflects the goal of the training process which is to train the model to make predictions as accurate as possible. For example, mean square error is defined as
\begin{align}
    \mathrm{MSE} = \sum_{i=1}^{n}(y_i-f(w;x_i))^2,
\end{align}
where $n$ is the number of samples, $w$ stands for the model parameters, $x_i$ is the feature vector of the $i$-th sample, $f(x_i)$ is the predicted label for the $i$-th sample, and $y_i$ is the true label of the $i$-th sample. Thus, in a regression problem using mean square error as the loss function, the model $w$ will get more punishment if the predicted label $f(x_i)$ is further from the true label $y_i$. The value of the loss function $L$ over $D_{train}$ is calculated as
\begin{align}
L(w;D_{train}) = \frac{1}{n_{train}} \sum_{i=1}^{n_{train}} L(w;d_i) = \frac{1}{n_{train}} \sum_{i=1}^{n_{train}} L(y_i, f(w;x_i)),
\end{align}
where $n_{train}$ is the number of samples in $D_{train}$. Here $L(w;D_{train})$ is also known as the empirical risk. The change in the value of $L(w;D_{train})$ shows the training progress of the model $w$. The lower the value of $L(w;D_{train})$ is, the better the model $w$ is trained on $D_{train}$. Note that norms should be added to the loss function to avoid the overfitting problem. In particular, overfitting means the model is too closely fit to $D_{train}$ and has a poor performance when dealing with new data.

\paragraph{\textbf{Training~(Optimization)}} After the loss is calculated, the model will be optimized iteratively to achieve a better generalization ability. This is also known as the training process. Typical optimization algorithms such as gradient decent are used to adjust parameters of the model according to the loss and the model structure. Considering that heavy calculation is required in this step, computational tricks like back-propagation are usually applied to improve system efficiency.

\paragraph{\textbf{Inference~(Prediction)}} With a trained model $f(w;\cdot)$, we can input a feature vector $x$ to it. Then, after some calculation, the model outputs a predicted label $f(w;x)$. This process is also known as the inference process. Usually inference is done by doing several times of matrix multiplication.

\subsection{Task Definition} \label{subsec:machine_learning_task_definition}

In machine learning, we first design the model structure $f$ according to what kind of problem we are going to solve. Then the goal is to find out a parameter vector $w$ which minimizes the expectation of the loss function. This can be expressed as
\begin{align}
\argmin_{w}\mathbf{E}[L(y, f(w;x))],
\end{align}
where $x \in X$ is the input feature and $y \in Y$ is the output label. It is assumed that the feature space $X$ and the label space $Y$ obey a joint probability distribution $P(x,y)$. Since the real $P(x,y)$ is unknown and is impossible to be figured out on most occasions, we are not able to directly optimize our model parameter $w$ to minimize the expectation of the loss function $\mathbf{E}[L(y, f(w;x))]$. As an approximation, we minimize the empirical risk $L(w;D_{train})$ on our training set $D_{train}$. This approximation requires $D_{train}$ to be a set which contains Independent and Identically Distributed (IID) random variables drawn from $P(x,y)$. Now the core learning task is expressed as
\begin{align}
\argmin_{w}L(w;D_{train})) = \argmin_{w}\frac{1}{n_{train}} \sum_{i=1}^{n_{train}} L(w;d_i) = \argmin_{w}\frac{1}{n_{train}} \sum_{i=1}^{n_{train}} L(y_i, f(w;x_i)),
\end{align}
where $n_{train}$ is the number of samples in $D_{train}$, $d_i = (x_i, y_i)$ is the $i$-th sample in $D_{train}$.

\subsection{Machine Learning Process} \label{subsec:machine_learning_process}
The procedure of machine learning can be divided into two parts: (1) Designing a suitable model structure based on the machine learning task; and (2) Examining the performance of the model and optimize it accordingly using large amounts of data.

\paragraph{\textbf{Model Design}} For the model design part, there already exists many well-designed models that have been proposed to solve different kinds of tasks, as shown in Fig.~\ref{fig:supervised_learning_taxonomy}. A good survey on popular traditional machine learning algorithms and models can be found in~\cite{Wu:2007:TAD:1327434.1327436}. What's more, in recent years, deep~learning~\cite{lecun2015deep} is proved to be very effective in many areas such as image recognition and natural language processing. The word ``deep'' is used to describe the multi-layer structure of the neural network used by it. The concept of deep learning can be further divided into Convolutional~Neural~Networks~(CNN), Recursive~Neural~Networks~(RNN), Generative~Adversarial~Networks~(GAN), and so on according to the neural network structure. Considering that the scale of a deep learning model can be very large, it is usually trained on high-performance servers. In general, deep learning is a powerful machine learning technique which is based on neural network and benefited from the increase in the amount of training data.


\paragraph{\textbf{Model Optimization}} Regarding the model optimization part, it is done through machine learning optimizers. Generally speaking, an optimizer focuses on how to optimize the model to reach its best performance based on the given dataset. For example, Gradient~Descent~(GD) calculates the gradient of the loss function and uses this gradient to optimize the model parameters. In each training iteration, according to the calculated gradient, the model parameters $w$ takes a step toward the optimal point where $L(w;D_{train})$ is minimized. However, the computation cost of GD is so high that it is not suitable for being applied to those models with a large number of parameters. Thus, many different schemes have been designed to find a better balance between the convergence speed and the computational cost. For those machine learning algorithms which are not suitable for using GD-based methods as their optimizers, they have their specific optimization algorithms, such as Sequential~Minimal~Optimization~(SMO) for SVM and Canopy for $k$-means.

\subsection{GD-Based Optimizers} \label{subsec:gdbased_optimizers}
Although there are many kinds of optimizers in centralized machine learning, in this survey, we mainly focus on the GD-based optimizers which originate from centralized machine learning and are now being widely used in distributed training and deep learning. In this section, we introduce several GD-based optimizers and analyze their advantages. We list the basic information of them in Table~\ref{tab:optimizer}. Note that the column ``Speed'' in the table has considered both the computational cost per iteration and the total number of iterations to reach convergence. Among these optimizers, SGD, SVRG, ADAM, and Hogwild! are designed for serial centralized machine learning or parallel multi-threads centralized machine learning. ASGD, EASGD, and DC-ASGD are actually designed for a {\em server-worker} scheme, which means they should be classified as distributed training techniques. However, we still choose to introduce them here for the convenience of comparing them with other optimizers and studying how GD-Based optimizers have developed from centralized to distributed.

\begin{table*}[!t]
  \renewcommand\arraystretch{1.3}
  \centering
  \caption{GD-based optimizers.}
  \label{tab:optimizer}
  \resizebox{\textwidth}{!}{
    \begin{tabular}[t]{|c|c|c|c|c|}
    \hline
    \textbf{Name} & \textbf{Mathematical Format} & \textbf{Convergence} & \textbf{Speed} & \textbf{Experimental Scale}\\
    \hline
    \textbf{SGD} & Gradient~(1 sample) & Sublinear~(all) & Baseline & An email dataset~(100 machines)~\cite{zinkevich2010parallelized}\\
    \hline
    \textbf{SVRG} & Gradient~(all samples) & \tabincell{c}{Linear~(strongly convex)\\ Sublinear~(convex)\\ Sublinear~(non-convex)} & Slow~$\downarrow$ & \tabincell{c}{MNIST~\cite{lecun1998MNIST} and CIFAR-10~\cite{hinton2007learningcifar10}~(1 machine)}\\
    \hline
    \textbf{ADAM} & SGD + Moment & Not Guaranteed & Fast~$\upuparrows$ & \tabincell{c}{MNIST and CIFAR-10~(1 machine)}\\
    \hline
    \textbf{Hogwild!} & SGD & \tabincell{c}{Linear~(strongly convex,\\ constant stepsize)} & Equivalent & \tabincell{c}{RCV1~\cite{lewis2004rcv1}, Netflix, KDD, Jumbo, DBLife,\\ and Abdomen~(10 machines)}\\
    \hline
    \textbf{ASGD} & SGD & Sublinear~(strongly convex) & Equivalent & A speech dataset and ImageNet~\cite{russakovsky2015imagenet}~(128 workers)\\
    \hline
    \textbf{EASGD} & SGD + Elastic Update & ?~(complicated form) & Fast~$\uparrow$ & CIFAR-10~(16 workers), ImageNet~(8 workers)\\
    \hline
    \textbf{DC-ASGD} & SGD + Delay Compensation & Sublinear~(strongly convex) & Fast~$\uparrow$ & CIFAR-10~(8 workers), ImageNet~(16 workers)\\
    \hline
    \end{tabular}%
    }
\end{table*}%

\paragraph{\textbf{SGD}} Stochastic Gradient Descent~(SGD)~\cite{robbins1951stochastic} was first proposed in 1951. The main advantage of SGD is that it greatly reduces the computational cost in each iteration compared with GD. Its core equations are very similar to those of GD and are shown as follows:
\begin{align}
    & g_t = \nabla L\left(w_{t-1}; d_i\right),
    & w_t = w_{t-1} - \eta \cdot g_t,
\end{align}
where $t$ is the timestamp for the current training iteration, $w_{t-1}$ is the model at time $t-1$, $g_t$ is the gradient of the loss function $L$ with model $w_{t-1}$ and a randomly selected data sample $d_i$, and $\eta$ is the learning rate. The feature that it only uses one sample to compute gradients in each iteration greatly reduces the computational cost. However, SGD should not be directly applied to client-based training since it cannot handle the bias caused by the non-IID local dataset (will~be~introduced~in~Section~\ref{subsec:client_training_challenges}). Modification to SGD is necessary to fit the scenario of client-based training~(\eg,~FedAvg~\cite{mcmahan2016communication}). Although the solution for client-based training on mobile devices is unlikely to choose the original SGD as the optimizer, we have to say that SGD still works well on many other occasions~(such as server-based distributed training) due to its low computational cost and high training efficiency.
    
\paragraph{\textbf{SVRG}} Stochastic~Variance~Reduced~Gradient~(SVRG)~\cite{johnson2013accelerating} aims at accelerating the convergence speed of SGD by applying noise reduction methods. Compared with GD, SGD does much less computation in each iteration but has a lower convergence speed. Bottou~\et~\cite{bottou2018optimization} discovered that one reason for this is the existence of noise in the estimate of the gradient, which can be also considered as the variance of gradients. Thus, SVRG uses the overall gradient to make corrections and thus reduces the noise. The core equations are shown as follows:
\begin{align} 
    & g_t' = \nabla f\left(w_{t-1}; d_i\right) - \nabla f\left(\bar{w}; d_i\right) + \bar{g},
    & w_t = w_{t-1} - \eta \cdot g_t',
\end{align}
where $\bar{w}$ is an averaged model which is updated every $k$ iterations and $\bar{g}$ is the gradient averaged among all data samples at point $\bar{w}$. The first term $\nabla f(w_{t-1}; d_i)$ on the right side of the first equation is exactly the $g_t$ used in SGD. The core idea of SVRG is to make the upper bound of gradients' variance keep reducing during training by using correction $(- \nabla f(\bar{w}; d_i) + \bar{g})$. McMahan~\et~\cite{mcmahan2016communication} found out that SVRG can cooperate well with some distributed optimization algorithms like DANE~\cite{shamir2014communication} by working as the local optimizer. The main factor which limits SVRG's applicability to client-based training is the high computational cost of periodically calculating the overall gradient $\bar{g}$.
    
\paragraph{\textbf{ADAM}} Adam is an optimization algorithm based on SGD aiming to accelerate the convergence speed by adaptively tuning the learning rate. It was proposed in 2014~\cite{kingma2014adam}. Before Adam, there already exists some algorithms trying to improve SGD through making use of the moment/momentum of gradients, such as SGD~with~Momentum~(SGDM)~\cite{qian1999momentum}, AdaGrad~\cite{duchi2011adaptive}, and RMSProp~\cite{RMSProp}. Adam combines the advantage of AdaGrad and RMSProp and uses both the first moment estimate and the second moment estimate. Its equations are given as follows:
\begin{align}
\begin{aligned}
    & m_t = \beta_1 \cdot m_{t-1} + \left(1-\beta_1\right) \cdot g_t,
    & v_t = \beta_2 \cdot v_{t-1} + \left(1-\beta_2\right) \cdot (g_t)^2,\\
    & \hat{m}_t = \frac{m_t}{1-(\beta_1)^t}, \quad\quad \hat{v}_t = \frac{v_t}{1-(\beta_2)^t},
    & w_t = w_{t-1} - \eta \cdot \frac{\hat{m}_t}{\sqrt{\hat{v}_t}+\epsilon}.
\end{aligned} 
\end{align}
Here, $m_t$ is the first moment estimate and $v_t$ is the second moment estimate. $\beta_1,\beta_2 \in [0,1)$ are exponential decay rates for the moment estimates. $\hat{m}_t$ and $\hat{v}_t$ are the bias-corrected version of the moment estimates. $\epsilon$ is a small constant used to prevent division by zero. Experiments show that Adam can accelerate training. Considering that the number of communication rounds may be limited in client-based training, an efficient optimizer which can speed up the training process may help a lot. However, Wilson~\et~\cite{NIPS2017_7003} and Reddi~\et~\cite{sashank2018convergence} discovered that on some special occasions, ADAM may fail to reach convergence. This can be the main barrier for applying ADAM to client-based training.

\paragraph{\textbf{Hogwild!}.} Hogwild!~\cite{recht2011hogwild} aims to prove that parallel SGD can be implemented without any locking. It shows that when the optimization problem is strongly convex and sparse, most updates only modify small subsets of all parameters, which means the whole update process can be run asynchronously without locking. However, it has only been tested on traditional machine learning problems like sparse SVM and matrix completion. Whether Hogwild! is suitable for complex tasks, such as deep learning and client-based training, still remains unknown.
    
\paragraph{\textbf{ASGD}} Asynchronous~SGD~(ASGD)~\cite{NIPS2012_4687} is a simple attempt for making use of more workers to train a huge deep network through asynchronous methods with SGD. Compared with synchronous SGD, ASGD won't suffer from the straggler problem which is a huge obstacle to deploying distributed machine learning on heterogeneous mobile devices. Since no waiting is needed, all workers can make best use of their resources and together accelerate the training. The problem is that in asynchronous methods, the delayed gradients may be unsuitable for being applied to the current updated model. The delay error can cause fluctuation in weights and have negative effects on the model according to~\cite{lian2015asynchronous, avron2015revisiting}. This delay error problem can be even worse in client-based training due to the large number of workers and the high frequency of model update.
    
\paragraph{\textbf{EASGD}} The purpose of Elastic~ASGD~(EASGD)~\cite{zhang2015deep} is to reduce the communication cost between workers and the parameter server during parallel training. Each worker's local model is not replaced by the global model in each communication round. The communication and coordination of work among all workers is controlled by an elastic force that links the local parameters with a center variable stored by the parameter server. The update rules are shown as follows:
\begin{align}
    & w_t^i = w_{t-1}^i - \eta \cdot \left(g_t^i + \rho \cdot \left(w_{t-1}^i - \bar{w}_{t-1}\right)\right),
    & \bar{w}_t = \bar{w}_{t-1} + \eta \cdot \sum_{i=1}^p \rho \cdot \left(w_{t-1}^i - \bar{w}_{t-1}\right),
\end{align}
where $i$ is a random index of a worker, $p$ is the number of workers, $\rho$ is the control parameter for the elasticity, and $\bar{w}_t$ is the center variable. The center variable $\bar{w}_t$ is updated as a moving average which is taken in both time and space over all local parameters. The elastic design allows workers to do more exploration in its nearby parameter space, which can do good to the model performance. The effectiveness of EASGD has only been analyzed for quadratic and strongly convex objectives. It is worthy to worry about that the model may become even worse if the elastic hyperparameter isn't set properly and causes the workers to explore too far away from the center variable.

\paragraph{\textbf{DC-ASGD}} Delay~Compensated~ASGD~(DC-ASGD)~\cite{zheng2017asynchronous} focuses on mitigating the error caused by delayed gradients in ASGD. The main idea of DC-ASGD is to use the first-order term in Taylor series to compensate for the delayed gradient and use an approximation of the Hessian Matrix to reduce the computational cost. At time $t+\tau$, to update model $w_{t+\tau}$, the original delayed gradient $g(w_t)$ for old model $w_t$ will be replaced by the delay-compensated gradient which is expressed as:
\begin{align}
g\left(w_t\right) + \lambda g\left(w_t\right) \odot g\left(w_t\right) \odot \left(w_{t+\tau} - w_t\right),
\end{align}
where $\lambda$ is a variance control parameter set by the server. The only additional information needed for compensation is the historical model $w_t$, which means this method is easy to implement. However, since experiments have been done with no more than 16 workers, DC-ASGD's performance under large numbers of workers still needs to be studied. In addition, another key problem of applying it to client-based training is that the server needs to spend large additional storage to store the historical models for all workers.

\subsection{Applicability of GD-Based Optimizers to Client-Based Training} \label{subsec:applicability_of_gdbased_optimizers}
In this section, we compare the applicability of the GD-based optimizers to client-based training in detail, as shown in Table~\ref{tab:optimizers_applicability}. The main difficulty of applying these optimizers to client-based training is that mobile devices may not have enough resources to run them on large models. A possible solution is to use distributed optimization algorithms~(will be introduced in Section~\ref{subsec:distributed_optimization_algorithms}) to decompose an original large problem into multiple small sub-problems. The aforementioned optimizers should cooperate with distributed optimization algorithms and work as the local optimizer. Therefore, we analyze the applicability of GD-based optimizers from two aspects: (1)~whether the {\em time and space complexities} are acceptable and affordable for mobile devices; and (2)~whether the optimizers support distributed computing in terms of {\em scalability, asynchronization, and delay solution}.

\paragraph{\textbf{Time Complexity}} For client-based training, optimizers with lower time complexity are preferred since they require fewer resources. Hard~\et~\cite{DBLP:journals/corr/abs-1811-03604} demonstrate that running SGD on mobile devices is feasible. We regard SGD as the baseline here. For SVRG, its periodical calculation of the overall gradient incurs huge local computation overhead. For ADAM, to accelerate the gradient descent step, it introduces additional gradient processing steps which contains several times of matrix multiplication. So ADAM's time complexity is relatively higher but still acceptable. For other GD-based optimizers, the local computation step is just SGD. Thus, regarding the time complexity, most of these GD-based optimizer~(except for SVRG) are applicable to client-based~training.

\paragraph{\textbf{Space Complexity}} Some optimizers use additional information to accelerate training or mitigate errors. This incurs additional space overhead. For SVRG, it keeps an overall gradient and a model to reduce the variance of gradients. For ADAM, it keeps the last gradient and the moment to accelerate training. For DC-ASGD, it needs to store a historical model for each worker for delay compensation. Therefore, these three optimizers require additional memory and storage space. For other GD-based optimizers, they do not need additional space. Therefore, regarding the space complexity, SGD, Hogwild!, ASGD, and EASGD are preferred.

\begin{table*}[!t]
  \renewcommand\arraystretch{1.3}
  \centering
  \caption{Applicability of GD-based machine learning optimizers to client-based training.}
  \label{tab:optimizers_applicability}
  \resizebox{0.92\textwidth}{!}{
    \begin{tabular}[t]{|c|c|c|c|c|c|c|c|}
    \hline
    \textbf{Name} & \textbf{Time Complexity} & \textbf{Space Complexity} & \textbf{Scalability} & \textbf{Asynchronization} & \textbf{Delay Solution} \\
    \hline
    \textbf{SGD} & Baseline \Checkmark& Basic Model \Checkmark& 100 machines \Checkmark& No \XSolidBrush& / \\
    \hline
    \textbf{SVRG} & High$\upuparrows$ \XSolidBrush& \tabincell{c}{Historical Gradient,\\ Historical Model \XSolidBrush} & 1 machine & No \XSolidBrush& / \\
    \hline
    \textbf{ADAM} & High$\uparrow$ & Historical Gradient \XSolidBrush& 1 machine & No \XSolidBrush& / \\
    \hline
    \textbf{Hogwild!} & Equivalent \Checkmark& Basic Model \Checkmark& 10 machines & Yes \Checkmark& None \XSolidBrush\\
    \hline
    \textbf{ASGD} & Equivalent \Checkmark& Basic Model \Checkmark& 128 workers \Checkmark& Yes \Checkmark& None \XSolidBrush\\
    \hline
    \textbf{EASGD} & Equivalent \Checkmark& Basic Model \Checkmark& 8 workers & Yes \Checkmark& Not Needed \Checkmark\\
    \hline
    \textbf{DC-ASGD} & Equivalent \Checkmark& Historical Model \XSolidBrush& 16 workers & Yes \Checkmark& Compensating \Checkmark\\
    \hline
    \end{tabular}%
    }
\end{table*}%

\paragraph{\textbf{Scalability}} We use ``machines'' to express that multiple machines cooperate with each other. We use ``workers'' to show that a group of workers are managed by a central server and may not need to communicate with each other. For synchronous optimizers, their scalability is limited by the time-consuming synchronization step. The ``thundering herd'' problem also add difficulty to implementing large-scale synchronous systems. For asynchronous optimizers, the obstacles are the update covering problem~(\ie,~later updates rewrite parameters and cover earlier updates.) and the delay error problem. With more workers, asynchronous training becomes less stable. However, the impact of these problems has not been clearly analyzed yet. Thus, we only list the number of machines/workers used in experiments to reflect the potential scalability. According to Table~\ref{tab:optimizers_applicability}, SGD and ASGD have shown good scalability.

\paragraph{\textbf{Asynchronization}} SGD, SVRG, and ADAM are designed for centralized machine learning with only one machine. They require a costly synchronization step to ensure convergence when being applied to parallel training. By allowing the asynchronous update of the global model, the time-consuming synchronization step can be avoided. For example, Hogwild!, ASGD, EASGD, and DC-ASGD are designed for asynchronous machine learning.

\paragraph{\textbf{Delay Solution}} The unstable network condition and the limited resources in client-based training can easily cause update delay. In asynchronous training scheme, the global model may have already been updated by others when the gradient arrives at the server. The delayed gradient thus becomes less accurate for the current global model. Directly applying delayed gradients to the global model can slow down the convergence speed due to the delay error. Hogwild! and ASGD have no solution for delayed gradients. They just ignore the delay error. EASGD does not need a delay solution because its global model is a moving average instead of being updated by gradients. DC-ASGD compensates delayed gradients by using Taylor series expansion. Therefore, regarding delay solution, EASGD and DC-ASGD outperform Hogwild! and ASGD.

%% file: 3_distributed_training.tex
\section{Server-Based Distributed Training} \label{sec:server_distributed_training}

Although centralized machine learning has shown good performance in many kinds of tasks, it cannot catch up with the growing demand of processing more data and training larger models. Under this circumstance, server-based distributed training techniques have been developed. The concept of server-based training means that operations and executions of the machine learning model are all done on servers. In this section, we first introduce the motivations and then discuss distributed parallelism categories and distributed optimization algorithms.

\subsection{Motivations} \label{subsec:distributed_training_motivations}

\paragraph{\textbf{Training Acceleration}} The original centralized machine learning scheme can only use the computation power of a single machine, which indicates that it may require a long time to train a good model when dealing with large amounts of data. A potential solution is to distribute data on different machines and let them process data samples simultaneously. As the heavy calculation is distributed to multiple machines and executed parallel, the training process is significantly~accelerated.

\paragraph{\textbf{Large Model Support}}
Large-scale deep learning has shown its effectiveness in many areas and has gained rapid development. However, for those large models that have billions and trillions of parameters to be optimized, the resources on a single machine can hardly support the learning task. Thus, researchers have studied the feasibility of dividing the large model into different parts and training them in parallel with multiple machines.

\subsection{Task Definition} \label{subsec:distributed_training_task_definition}

In Section~\ref{subsec:machine_learning_task_definition}, we have mentioned that the objective function is: $\argmin_{w}L(w;D_{train}))$. There are three key components in the formula: (1) Loss Function $L$; (2) Model parameters $w$; and (3) Dataset $D_{train}$. For these three components, we determine which of them to be divided:
\begin{itemize}
    \item Loss Function $L$: Since $L$ usually takes only a little storage, it is not necessary to divide it.
    \item Dataset $D_{train}$: If $D_{train}$ is large, we can divide and store it on several machines. After machines generate model updates through local training, the updates are transferred and aggregated to generate a new global model. This is known as distributed training with data parallelism.
    \item Model $w$: If $w$ is large and contains huge amounts of parameters, we can let each machine process only one part of $w$. Intermediate results over parts of the model are transferred between machines. This is known as distributed training with model parallelism.
\end{itemize}
Note that both data parallelism and model parallelism let the whole dataset flow through the whole model, which guarantees the effectiveness of training. We normally do not divide the dataset and the model at the same time because this may cause incomplete training and result in performance~loss.

\subsection{Parallel Training Categories}\label{subsec:parallelism_categories}

\begin{figure}[t]
	\centering
	\subfloat[Model Parallelism]{\label{fig:parallelism:model}
	\begin{minipage}{.31\linewidth}
	\centering
	\includegraphics[width=\textwidth]{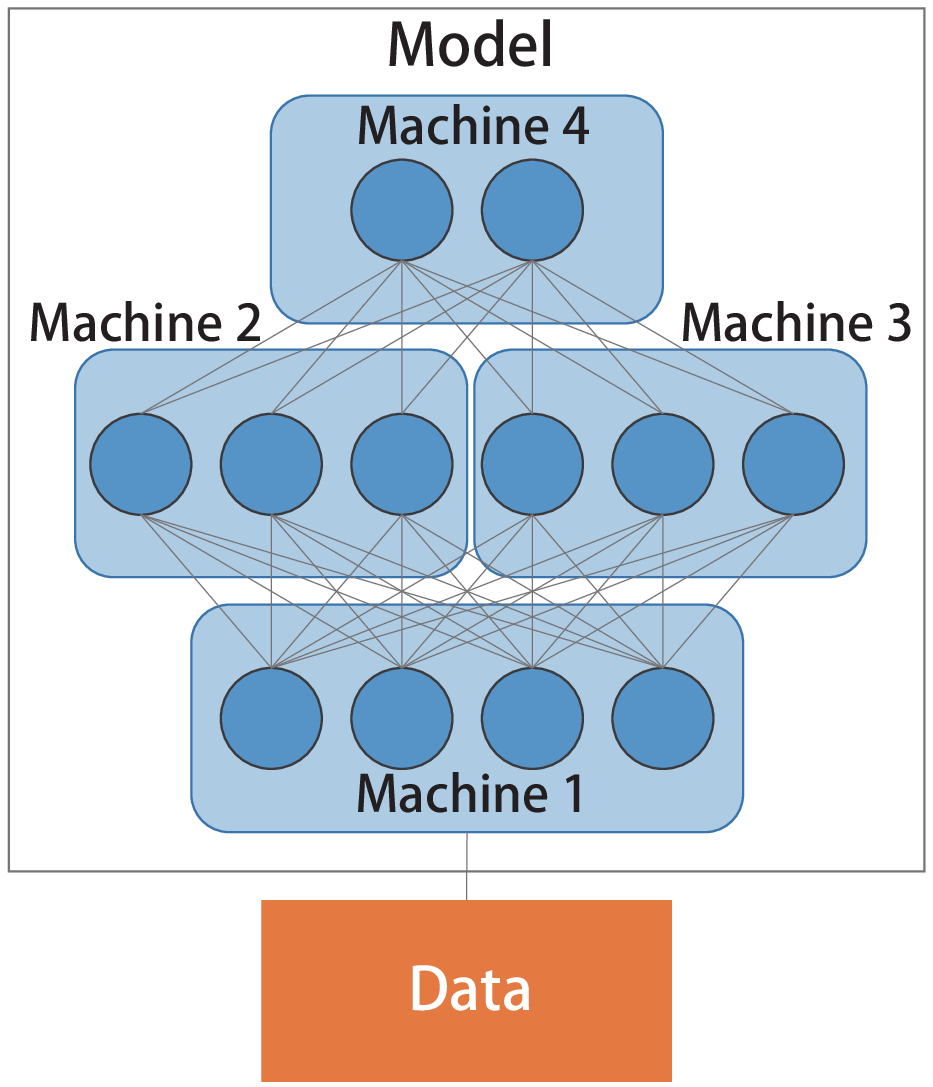}
	\end{minipage}
	}
	\subfloat[Data Parallelism]{\label{fig:parallelism:data}
	\begin{minipage}{.62\linewidth}
	\centering
	\includegraphics[width=\textwidth]{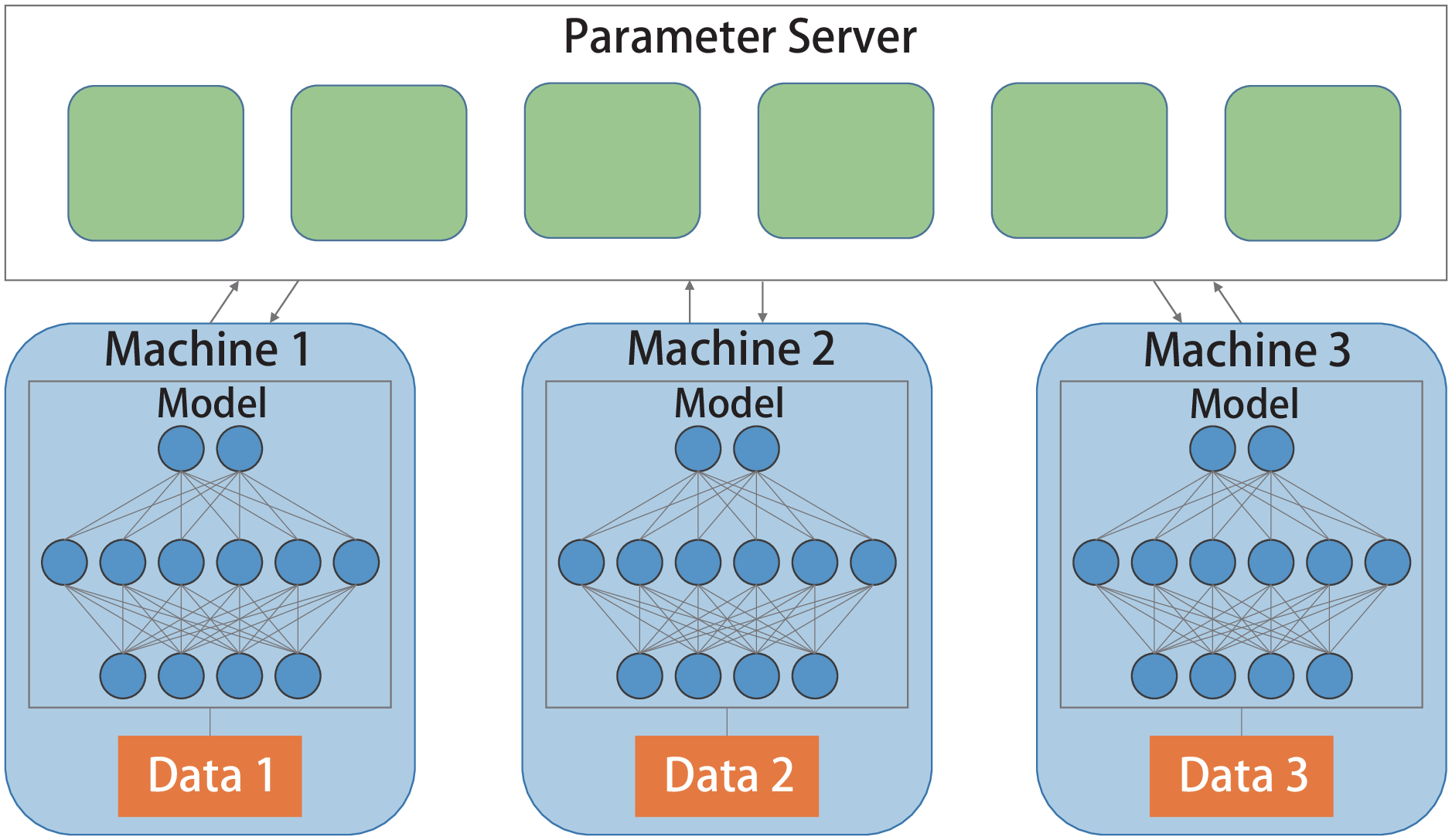}
	\end{minipage}
	}
	\caption{Architectures of model parallelism and data parallelism for server-based distributed training.}\label{fig:parallelism}
\end{figure}

As shown in Fig.~\ref{fig:parallelism}, there are two complementary architectures for distributed machine learning, namely data parallelism and model parallelism.

\paragraph{\textbf{Data Parallelism}} In most occasions, a large dataset that contains various kinds of data samples is very helpful for training a well-performed model. Sometimes the dataset is so large that it cannot be stored on a single machine. It is also possible that the huge amount of data results in prohibitively slow training process. According to~\cite{peteiro2013survey}, distributed machine learning with data parallelism has emerged to solve the data storage problem and accelerate the training. In this scheme, the whole dataset is divided into sub-sets and distributed on machines. Each machine keeps a copy of the model and trains it based on the locally available part of data. After several iterations of training, the local models may become quite different from each other. The information is gathered to generate an updated global model. This process is called data aggregation. Then, if the performance of the new global model is still not satisfying, another round of training is started. With data parallelism, more data is processed simultaneously, which means it speeds up the training.
    
\paragraph{\textbf{Model Parallelism}} In some special machine learning tasks, the model can be so large that it is too slow and even not able to be trained and run on a single machine. This problem is particularly serious in deep learning tasks. Thus, large-scale distributed deep networks are proposed in~\cite{NIPS2012_4687} trying to deal with it. Model parallelism methods are adopted and used to train large models with billions of parameters. In this scheme, each machine keeps a small part of the whole model. During training, the data flows through machines in order to be processed by the local sub-models. On most occasions, every round of training needs the cooperation of all machines. Therefore, this process should be done sequentially since the inputs of some machines depend on the outputs of others. In this situation, using a scheduler to manage the training process may be helpful for solving the dependency between machines. Compared with data parallelism, model parallelism is more complex and also harder to implement due to the strong cooperation among machines.

\subsection{Parallel Communication Frameworks} \label{subsec:parallelization_frameworks}
Parallel communication frameworks help realize parallel training. Without a well-designed communication scheme, the limitation on the network bandwidth may become a troublesome bottleneck for the whole system. The frameworks can be divided into the following three kinds: (1)~\textbf{MapReduce/AllReduce}; (2)~\textbf{Parameter Server}; and (3)~\textbf{Data Flow}. We briefly introduce their design in Section \ref{subsubsec:mapreduce} - \ref{subsubsec:data_flow}. Then we give a simple discussion on this topic in Section \ref{subsubsec:parallel_framework_discussion}.

\subsubsection{MapReduce/AllReduce} \label{subsubsec:mapreduce}

\begin{figure}
    \centering
    \includegraphics[width=0.8\columnwidth]{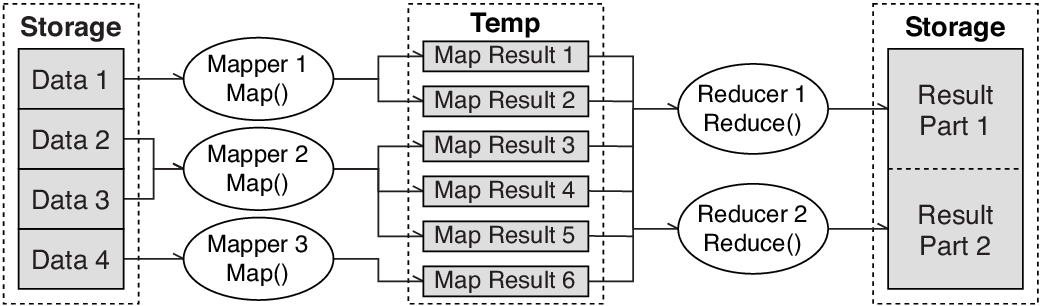}
    \caption{Process of MapReduce.}
    \label{fig:mapreduce}
\end{figure}

\paragraph{\textbf{Design of MapReduce}} In MapReduce, the map operation distributes data and tasks to workers and the reduce operation aggregate all results. The general process of MapReduce is shown in Fig.~\ref{fig:mapreduce}. To accomplish the distributed computation task, several mappers and reducers are set on available nodes. Mappers read data from the storage, perform the mapping in parallel, and generate intermediate results. Reducers then aggregate intermediate results and generate the final result.

\begin{figure}
    \centering
    \includegraphics[width=0.84\columnwidth]{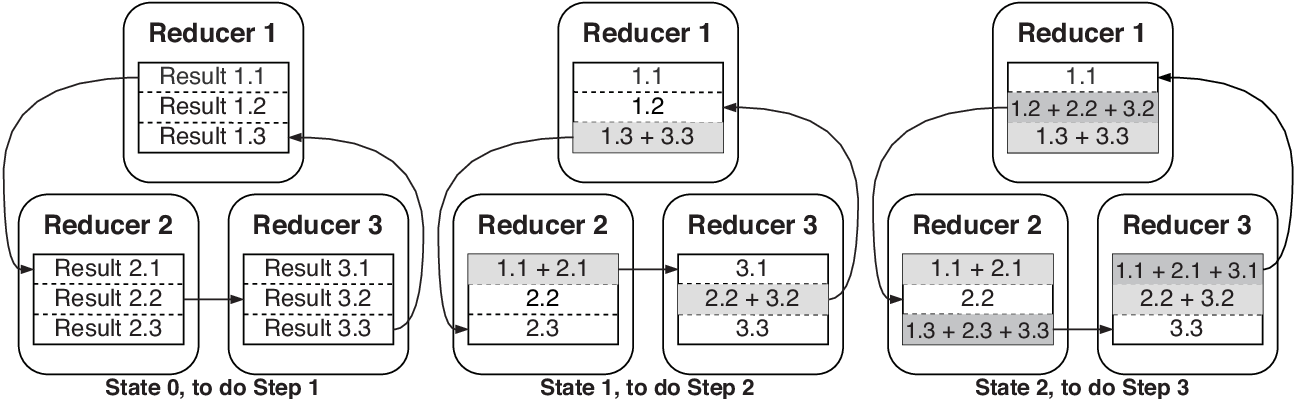}
    \caption{Process of Ring AllReduce.}
    \label{fig:allreduce}
\end{figure}

\paragraph{\textbf{Design of AllReduce}} One problem of MapReduce is that it takes huge communication costs to transfer intermediate results to the reducers. To deal with this problem, AllReduce is proposed and integrated into Message~Passing~Interface~(MPI). In AllReduce, all worker nodes also work as reducers. Parts of the intermediate results are transferred between workers if necessary. With this design, the amount of transferred data decreases. The network burden is also relieved since all the workers' bandwidths are used during reducing. AllReduce has been realized using many different topologies~\cite{patarasuk2009bandwidth}. Here we use ring topology~(Fig.~\ref{fig:allreduce}) as an example to explain AllReduce. For $k$~(here $k=3$) nodes, we first use $(k-1)$ steps to make each node keeps $\frac{1}{k}$ of the final aggregated result. Then with another $(k-1)$ ring communication steps, the result on each node can be completed.

\paragraph{\textbf{Pros/Cons}} In machine learning, MapReduce/AllReduce is commonly used to make multiple machines cooperate with each other. One advantage of MapReduce/AllReduce is that it is easy to deploy. MapReduce has been applied to server-based distributed training in~\cite{meng2016mllib, 5767930}. Here, mapping is to retrieve the global model and generate model updates. Reducing is to aggregate model updates and update the global model. Baidu brought Ring AllReduce technique to deep learning in 2017~\cite{baidu-allreduce}. AllReduce has now been supported in TensorFlow. In addition, we can use a third-party library, called Horovod~\cite{sergeev2018horovod}, to simplify the implementation of AllReduce in deep learning. One problem of MapReduce/AllReduce is that it can be easily blocked by stragglers as synchronization is needed.
    
\subsubsection{Parameter Server} \label{subsubsec:parameter_server}

\paragraph{\textbf{Design of Parameter Server}}
The parameter server~(PS) can be either a single server or a server cluster that takes charge of the task arrangement but doesn't do the task by itself. The tasks are actually done by the workers. Each worker only has to communicate with the central server to pull or push data and has no need to be aware of other participators. This means one worker's computation task is independent of others' so that asynchronous working is possible. PS is more robust than MapReduce/AllReduce since it supports asynchronous communication and thus won't suffer from the straggler problem. An example of PS has been shown in Fig.~\ref{fig:parallelism:data}.

\paragraph{\textbf{Development of Parameter Server}}
The idea of PS came from the parallel Latent~Dirichlet~Allocation~(LDA) architecture~\cite{smola2010architecture}. This first-generation PS used Memcached as the storage of parameters and managed the synchronization of workers. The lack of flexibility and performance is its main disadvantage. After that, YahooLDA~\cite{ahmed2012scalable} and Distbelief~\cite{NIPS2012_4687} followed this idea and improved PS's design for specific applications. Petuum~\cite{ho2013more} was a more general platform based on YahooLDA but it placed more constraints on worker threading models. These works are all considered as the second-generation PS. To build a more robust system, the third-generation PS was proposed and implemented in~\cite{186214}. PS has been applied to distributed deep learning tasks by Google in~\cite{NIPS2012_4687}.

\paragraph{\textbf{Pros/Cons}}
Here, we first introduce the advantages of the third-generation PS: (1)~{\em Efficient Communication:} Communication has been optimized for learning tasks to reduce overhead, and asynchronous communication is supported; (2)~{\em Flexible Consistency:} The system allows three consistency models, including sequential consistency, eventual consistency, and bounded delay consistency; (3)~{\em Elastic Scalability:} New servers can join without rebooting the whole system; and (4)~{\em Fault Tolerance and Durability:} Chain replication is used to backup data entries on servers. The vector clock design enables a failed node to be quickly recovered to its original working status. Compared to AllReduce, one disadvantage of PS is that all transferred data must pass through the server. This can bring huge communication burden to the server. On the other side, the support for asynchronous training and the elastic scalability are great advantages of PS. Although the PS architecture is originally designed for server-based machine learning, some of its features such as efficient communication are also strongly needed in client-based machine learning.
    
\subsubsection{Data Flow} \label{subsubsec:data_flow}

\paragraph{\textbf{Design of Data Flow}} For distributed machine learning with model parallelism, a specially designed scheme called data flow can be applied. Unlike the above-mentioned two schemes in which each node has similar functions for the whole task, in data flow, different parts of the model are distributed on different machines, so their jobs vary from one to another. The whole computation process is organized using a directed acyclic graph. Nodes are units of the model, and edges describe how data flows. If data flows between two units which are stored on different machines, communication will be taken place. An example of this scheme has been shown in Fig.~\ref{fig:parallelism:model}.

\paragraph{\textbf{Pros/Cons}} The disadvantage of this scheme is that the failure of any machine can cause the graph to be incomplete and the system can no longer run. If we use redundancy to solve this problem, backups for every machine is necessary. This may result in an expensive cost. Thus, the data flow scheme is more suitable for the cooperation among several powerful and stable machines.

\subsubsection{Discussion} \label{subsubsec:parallel_framework_discussion}
From these techniques, we can see that server-based distributed training focuses on the cooperation between powerful machines. Its core idea is using data parallelism or model parallelism to solve large-scale learning tasks. The different communication schemes are designed for dealing with the low bandwidth of the local area network compared with the shared memory. However, in client-based training, the challenges are mainly due to the limited resources and the unstable network, which is quite different from those of server-based distributed training. Even if some techniques such as data parallelism and parameter server can be referred to for designing client-based training, we still have to pay attention to the special limitations in its scenario.

\subsection{Distributed Optimization Algorithms} \label{subsec:distributed_optimization_algorithms}

This kind of algorithms focus on how to better manage a large scale of workers and how to use their resources to solve a huge complex task in a distributed way. A typical solution is to transform the original hard problem into much easier sub-problems. In this section, we introduce several distributed optimization algorithms. We list their key features in Table~\ref{tab:distributed_opyimization}.

\begin{table*}[!t]
  \renewcommand\arraystretch{1.3}
  \centering
  \caption{Distributed optimization algorithms for server-based distributed training.}
  \label{tab:distributed_opyimization}
  \resizebox{\textwidth}{!}{
    \begin{tabular}[t]{|c|c|c|c|c|}
    \hline
    \textbf{Name} & \textbf{Math Tool} & \textbf{Objective} & \textbf{Convergence Speed} & \textbf{Experimental Scale}\\
    \hline
    \textbf{ADMM} & \tabincell{c}{Dual Decomposition, \\ Method of Multipliers} & \tabincell{c}{Closed, Proper,\\ Convex} & Reach optimal as iteration number $k \rightarrow \infty$. & \tabincell{c}{Tested on 30GB data\\ with 80 workers.}\\
    \hline
    \textbf{DANE} & \tabincell{c}{Approximate\\ Newton Method} & \tabincell{c}{Strongly Convex,\\ Smooth} & \tabincell{c}{Linear convergence\\ for quadratic objectives.} & \tabincell{c}{Tested on CoverType~\cite{Dua:2019}, MNIST,\\ and ASTRO-PH~\cite{joachims2006training} with 64 workers.}\\
    \hline
    \textbf{CoCoA} & \tabincell{c}{Use dual variables to\\ do efficient merge.} & \tabincell{c}{Convex,\\ Smooth} & \tabincell{c}{Convergence rate is $\frac{m-1+\Theta}{m}$.\\ ($m$: number of workers, $\Theta$: convergence\\ rate of the local optimization method)} & \tabincell{c}{CoverType with 4 workers.\\ RCV1 with 8 workers.\\ ImageNet with 32 workers.}\\
    \hline
    \textbf{CoCoA+} & \tabincell{c}{Use dual variables to do\\ efficient additive merge.} & Convex & \tabincell{c}{Linear convergence for smooth\\ convex objectives. Independent\\ of the number of workers.} & \tabincell{c}{CoverType with 16 workers.\\ RCV1 with 16 workers.\\ Epsilon with 100 workers.}\\
    \hline
    \textbf{DiSCO} & \tabincell{c}{Inexact Damped\\ Newton Method} & \tabincell{c}{Strongly Convex,\\ Smooth,\\ Self-Concordant} & \tabincell{c}{Linear convergence.\\ Communication rounds $t \approx O((md)^{\frac{1}{4}}\log(\frac{1}{\epsilon}))$.\\ ($m$: number of workers, $d$: number of features)} & \tabincell{c}{Tested on CoverType, RCV1,\\ and News20~\cite{news20} with 64 workers.}\\
    \hline
    \textbf{Hydra} & \tabincell{c}{Randomized Coordinate\\ Descent Method} & \tabincell{c}{Strongly Convex,\\ Smooth} &  \tabincell{c}{Roughly linear convergence.\\ Reach $\epsilon$-accurate with possibility at least\\ ($1 - \rho$) after $O(\frac{d}{m}\log(\frac{1}{\epsilon \rho}))$ iterations.\\ ($m$: number of workers, $d$: number of features)} & \tabincell{c}{Speedup tested on 3TB data with\\ 512 workers. Convergence tested on\\ ASTRO-PH with 32 workers.}\\
    \hline
    \end{tabular}%
    }
\end{table*}%

\paragraph{\textbf{ADMM}} Alternating~Direction~Method~of~Multipliers~(ADMM)~\cite{boyd2011distributed} makes use of both the decomposability of Dual Ascent and the superior convergence properties of Method of Multipliers. Machines distributedly use augmented Lagrangian methods to solve local sub-problems based on local data and alternately compute some shared variables to solve the global problem. The idea of this algorithm can be traced back to the mid-1970s and has been used in distributed SVM training~\cite{forero2010consensus} in 2010. Although ADMM may take a large number of iterations to converge to high accuracy, it can usually reach modest accuracy within tens of iterations in practice.
    
\paragraph{\textbf{DANE}} Distributed~Approximate~NEwton~(DANE)~\cite{shamir2014communication} is an approximate Newton-like method. In every iteration, each worker separately takes an approximate Newton step with implicitly using its local Hessian and makes two rounds of communication. In contrast to ADMM, DANE can benefit from the fact that sub-problems are often similar in applications of machine learning. DANE performs well on smooth and strongly convex problems. It is proved that DANE can achieve linear convergence if the learning rate is close to 1 and the approximation for Hessian is good.

\paragraph{\textbf{CoCoA}} Communication-efficient distributed dual Coordinate~Ascent~(CoCoA) was first proposed in 2014~\cite{jaggi2014communication}. By making use of convex duality, after dividing the task into approximate local sub-problems, we can choose whether to solve the primal sub-problem or the dual problem. The main advantage of CoCoA is its flexibility. It allows machines to choose an local optimization method and train the model to arbitrary accuracy. What's more, the trade-off between local computation and communication can also be tuned easily by setting a specific parameter. Experiments show that CoCoA can achieve up to a $50\times$ speedup when dealing with problems like SVM, logistic regression, and lasso. Note that CoCoA~\cite{jaggi2014communication} was improved to a more general case CoCoA'~\cite{smith2017cocoa} in 2017~(CoCoA~\cite{jaggi2014communication} and CoCoA+~\cite{ma2015adding} are predecessors of CoCoA'~\cite{smith2017cocoa}.). The convergence for non-smooth or non-strongly convex objectives has been analyzed for CoCoA'.

\paragraph{\textbf{CoCoA+}} CoCoA+~\cite{ma2015adding} also makes use of the primal-dual problem to get optimization. Compared with CoCoA, CoCoA+ additionally studies and proves the convergence on non-smooth loss functions. Linear convergence is proved for convex smooth objectives, while sub-linear convergence is proved for convex non-smooth objectives. Furthermore, to get rid of the slowdown caused by averaging updates, CoCoA+ choose to add all updates. Experiments show that CoCoA+ only slows down a little as the number of workers increases, and it is faster than CoCoA for a large number of workers.
    
\paragraph{\textbf{DiSCO}} Distributed~Self-Concordant~Optimization~(DiSCO)~\cite{zhang2015disco} is a Newton-type method. Compared with DANE, DISCO uses a distributed preconditioned conjugate gradient method to compute inexact Newton steps in each iteration and gets a superior communication efficiency. One significant advantage of DiSCO is that compared with other algorithms, it has fewer parameters to be paid attention and adjusted. According to the experiments, when the number of workers increases to 16 and 64, DiSCO significantly outperforms ADMM and DANE on the convergence speed. DiSCO-S~\cite{ma2016distributed} is an improved version of DiSCO. It uses an approximated Hessian as its preconditioning matrix and uses Woodbury Formula to deal with the linear system more efficiently.

\paragraph{\textbf{Hydra}} HYbriD~cooRdinAte~(Hydra)~\cite{richtarik2016distributed} is a randomized coordinate descent method. This kind of methods are becoming popular in many learning tasks such as boosting and large-scale regression. In Hydra's design, the original data are partitioned and assigned to one node from the cluster of machines. Each node independently updates a random subset of its data based on a designed closed-form formula in each iteration. The updates are all parallelized.

\subsection{Applicability of Distributed Optimization Algorithms to Client-Based Training} \label{subsec:distributed_optimization_applicability}

Compared with server-based distributed training, client-based distributed training is harder to implement due to the large number of workers and the relatively limited resources. Thus, some distributed optimization algorithms designed for server-based training may no longer be applicable to client-based training. In what follows, we discuss the applicability of existing distributed optimization algorithms to client-based training. As shown in Table \ref{tab:distributed_optimization_applicability}, we investigate applicability mainly from three aspects: (1)~\textbf{Local Complexity}; (2)~\textbf{Communication Overhead}; and (3)~\textbf{Scalability}. They are key concerns in the context of client-based training. We also list some other pros and cons of these distributed optimization algorithms.

\begin{table*}[!t]
  \renewcommand\arraystretch{1.3}
  \centering
  \caption{Applicability of distributed optimization algorithms to client-based training.}
  \label{tab:distributed_optimization_applicability}
  \resizebox{\textwidth}{!}{
    \begin{tabular}[t]{|c|c|c|c|c|}
    \hline
    \textbf{Name} & \textbf{\tabincell{c}{Local Complexity}} & \textbf{\tabincell{c}{Comm. Overhead\\ (per worker per round)}} & \textbf{Scalability} & \textbf{Other Pro(s)/Con(s)}\\
    \hline
    \textbf{ADMM} & \tabincell{c}{A serially solvable\\ convex problem.} & AllReduce(model) & \tabincell{c}{Not theoretically analyzed.\\ Perform well in experiments.} & \tabincell{c}{Tested with numerical experiments.\\ Implemented in C using MPI~\cite{admmcode}. \Checkmark}\\
    \hline
    \textbf{DANE} & Do mirror descent. & 3 $\cdot$ Sizeof(model) & \tabincell{c}{Convergence rate is independent\\ of the number of workers. \Checkmark} & \tabincell{c}{Perform well only for\\ quadratic objectives. \XSolidBrush}\\
    \hline
    \textbf{CoCoA} & \tabincell{c}{Depend on local dual\\ optimization method.} & 2 $\cdot$ Numof(features) \Checkmark & \tabincell{c}{Slow down as the number\\ of workers grows. \XSolidBrush} & \tabincell{c}{Allow steering the trade-off between\\ communication and local computation. \Checkmark}\\
    \hline
    \textbf{CoCoA+} & \tabincell{c}{Depend on local dual\\ optimization method.} & 2 $\cdot$ Numof(features) \Checkmark & \tabincell{c}{Convergence rate is independent\\ of the number of workers. \Checkmark} & \tabincell{c}{Allow steering the trade-off between\\ communication and local computation. \Checkmark}\\
    \hline
    \textbf{DiSCO} & \tabincell{c}{Compute gradients\\ and Hessians.} & Numof(features)$^2$ \Checkmark & \tabincell{c}{Slow down as the number\\ of workers grows. \XSolidBrush} & \tabincell{c}{Require self-concordant objectives.\XSolidBrush}\\
    \hline
    \textbf{Hydra} & Do coordinate decent. & Not clearly analyzed. \XSolidBrush & \tabincell{c}{Speed up as the number\\ of workers grows.} \Checkmark & \tabincell{c}{Partition dataset by features,\\ require redistribution of data. \XSolidBrush}\\
    \hline
    \end{tabular}%
    }
\end{table*}%

\paragraph{\textbf{Local Complexity}}
Limited by the relatively poor computation resources~(CPU and memory) available on mobile devices, client-based training is very sensitive to the local complexity of algorithms. However, in server-based training, local complexity has not been paid much attention as the worker nodes are all regarded as powerful machines. We can hardly find accurate local complexity information in the above-mentioned distributed optimization work. Here, in Table~\ref{tab:distributed_optimization_applicability}, we just list the local computation methods which may reflect the local complexity to some degree.

In ADMM, DANE, DiSCO, and Hydra, the local optimization step is fixed. However, CoCoA and CoCoA+ allow multiple choices on the local solver~(Please refer to~\cite{smith2017cocoa} for detailed suggestions.), which means we can choose an appropriate local solver according to workers' available resources. This design is useful in client-based training since it improves the algorithm's compatibility for heterogeneous mobile devices. Thus, regarding the local complexity, CoCoA and CoCoA+ outperform the others.

\paragraph{\textbf{Communication Overhead}}
Communication overhead is another important issue in client-based training. Considering the complex network condition of mobile devices, high communication overhead not only brings expensive data transfer cost, but also increases the risk of transfer failure. Communication overhead can be further divided into the number of communication rounds and the amount of data transferred in each round.
The number of communication rounds is reflected by the convergence speed listed in Table~\ref{tab:distributed_opyimization}. Except for ADMM which requires large number of rounds to converge, the other algorithms can all achieve linear convergence.

The amount of transferred data per worker per round is listed in the Communication Overhead column in Table \ref{tab:distributed_optimization_applicability}. In ADMM, the communication step is accomplished by using AllReduce to update the model parameters. In DANE, the amount of data transferred per round is triple as much as the size of the model. Since the size of the model is not accurately provided, we cannot get the accurate communication overhead of ADMM and DANE. In CoCoA and CoCoA+, the amount of data transferred by each worker in each round is twice as much as the number of the features. They are quite communication-efficient when dealing with tasks with small feature space. In DiSCO, the amount of data transferred in each round is square as much as the number of features, which may also be acceptable. Although several communication schemes have been provided in Hydra, the amount of transferred data has not been clearly analyzed. Regarding communication overhead, CoCoA, CoCoA+, and DiSCO outperform the other algorithms.

\paragraph{\textbf{Scalability}}
Good scalability means we can assign the task to more workers and make use of more parallel resources. We compare the scalability of distributed optimization algorithms by checking if their convergence speed is affected by the number of workers. Results are listed in~Table~\ref{tab:distributed_optimization_applicability}.

In ADMM, although the scaled version of the algorithm is given, the scalability is only discussed for the scale of the datasets. Whether the scale of workers has an impact on ADMM is not theoretically analyzed. We could only say that the performance of ADMM is not greatly changed as the number of workers grows according to the comparison experiments between DANE and ADMM done in \cite{shamir2014communication}. In DiSCO and CoCoA, according to Table \ref{tab:distributed_opyimization}, the increase in the number of workers causes the slowdown of the convergence speed. This indicates that the scalability of DiSCO and CoCoA is not so good. In DANE and CoCoA+, the convergence rate is independent of the number of workers. In Hydra, according to Table \ref{tab:distributed_opyimization}, since it can reach $\epsilon$-accurate with possibility at least ($1 - \rho$) after $O(\frac{d}{m}\log(\frac{1}{\epsilon \rho}))$ iterations, the number of iterations actually decreases as the number of workers $m$ increases. This means Hydra speeds up as the scale of workers grows. Regarding scalability, DANE, CoCoA+, and Hydra outperform the other algorithms.

\paragraph{\textbf{Other Pros/Cons}}
Besides the above three key aspects which are mostly concerned by client-based training, these distributed algorithms also have some other pros/cons that should be incorporated for evaluating their applicability to client-based training. Details are presented in~Table~\ref{tab:distributed_optimization_applicability}.

DANE and DiSCO require strict assumptions on the objective function. Hydra partitions training data by features, which means it may require a data redistribution step. Thus, DANE, DiSCO, and Hydra have additional cons. ADMM has been implemented using MPI in~\cite{admmcode} and thus is easier to be applied. CoCoA and CoCoA+ allow steering the trade-off between communication and local computation, which gives us the freedom to change the training strategy according to available resources. For this part, we could say that ADMM, CoCoA, and CoCoA+ outperform the others.

Another problem is that all these distributed optimization algorithms require a synchronous update aggregation process in each communication round. Since the heterogeneity of mobile devices can easily cause the straggler problem, the synchronization may significantly degrade the system efficiency. This disadvantage should be carefully considered when applying distributed optimization algorithms to client-based training. Considering that some optimizers introduced in Section~\ref{subsec:gdbased_optimizers} have already supported the asynchronous scheme, we can refer to them and design new asynchronous distributed optimization algorithms for client-based training.

%% file: 4_client_inference.tex
\section{Client-Based Inference} \label{sec:client_inference}
In contrast to server-based machine learning where all operations over the model (including training and inference) are conducted on servers, client-based machine learning intends to pull down some computation tasks to clients so that the quality of service can be improved in terms of response latency, personalization, security privacy, and so on. Based on the execution phase, we divide client-based machine learning into client-based inference and client-based training. Client-based inference focuses on the local execution of trained machine learning models. In this section, we discuss the necessity and the feasibility of client-based inference.

\subsection{Motivations} \label{subsec:client_based_inference_motivations}

Motivations for client-based inference can be concluded as follows: (1) Client-based inference can reduce service latency and preserve user privacy; (2) Client-based inference helps lower the cost of cloud platforms for service producers; (3) It is feasible to deploy client-based inference thanks to the rapid development of mobile chipsets; and (4) It is convenient to deploy client-based inference using the off-the-shelf mobile machine learning frameworks.

\subsection{Challenges} \label{subsec:client_based_inference_challenges} One major challenge is that {\bf (1) inference tasks using complex or large-scale models can hardly be solved on mobile devices.} Even though the power of mobile chipsets has grown fast, it is still far away from that of high-performance servers. The CPU power and the memory capacity of mobile devices may have trouble supporting complex models, especially deep learning models.

Another challenge is that {\bf (2) the trade-off between inference accuracy and resource consumption needs to be carefully decided.} This is a fine-grained resource control problem. As client-based inference aims at improving the quality of service, we should never let client-based inference take too much resources or drain the battery. Considering that multiple models may be run simultaneously to provide different services, the trade-off between inference accuracy and resource consumption must be carefully decided to prevent resource contention.

\subsection{Current Advances} \label{subsec:client_based_inference_current_advances}
In the past few years, deep learning has become the main trend of machine learning. Mobile phones, as the most popular device in people's daily life, becomes the most promising platform for deep learning. Ravi~\cite{pmlr-v97-ravi19a} has proposed a method to generate compact on-device neural networks, which brings convenience to the deployment of client-based deep learning inference. Deep learning based apps such as~\cite{10.1145/2971648.2971731, 10.1145/2750858.2804262, Zeng:2017:MSM:3081333.3081336} have been built. General mobile deep learning frameworks~\cite{TensorFlowLite, PyTorch, alibaba2020mnn, NCNN, PaddleLite, CoreML} have been developed. Here, we introduce several mobile deep learning applications to show the current development of client-based inference.

\paragraph{\textbf{Mobile Computer Vision (CV)}} According to~\cite{Xu2019afirstlook}, photo beautify and face recognition together cover over a half of all mobile deep learning applications. For CV applications, server-based inference can bring high privacy risk as the raw sensitive images and videos have to be transferred through network. To solve this problem, DeepEye~\cite{10.1145/3081333.3081359} and DeepMon~\cite{10.1145/3081333.3081360} were proposed to support the on-device execution of deep vision models. Another problem is that the huge latency incurred by server-based inference may be unacceptable for applications such as mobile augmented reality. As a solution, DeepDecision~\cite{8485905} designs an on-device small deep learning model tiny-YOLO and automatically decides the inference to be done locally or remotely. By using optional client-based inference, the real-time capability of the system is guaranteed.

\paragraph{\textbf{Mobile Natural Language Processing (NLP)}} Another hot topic is to solve NLP tasks such as sentiment analysis, translation, and question answering on mobile devices. To better preserve users' privacy as well as reduce the service latency, DeQA~\cite{10.1145/3307334.3326071} adapts existing question answering systems to suit on-device running. Client-based inference is also applied to voice assistant and voice input applications to ensure that they can still function normally even the device is offline. Siri, an voice assistant developed by Apple, uses on-device deep learning~\cite{Capes2017, team2017deep} to improve the text-to-speech synthesis process. Google makes use of client-based inference to realize personalized speech recognition~\cite{7472820}. Georgiev~\et~\cite{10.1145/3081333.3081358} use mobile GPU to accelerate mobile audio sensing.

\paragraph{\textbf{Other Applications}} Client-based inference also plays an important role in other applications that require real-time responses or involve sensitive data processing. Liu~\et~\cite{10.1145/3090082} designed UbiEar to do smartphone-based acoustic event sensing and notification. This work greatly improves the quality of life for Deaf~or~Hard-of-Hearing~(DHH) people. Fang~\et~\cite{10.1145/3241539.3241559} designed NestDNN to dynamically select the structure of on-device deep learning CV models based on the trade-off between accuracy and current available resources. In addition to these published works, client-based inference are also applied to various kinds of applications (\eg, recommendation, movement tracking, and identity recognition) without being noticed~\cite{Xu2019afirstlook}. It is certain that people's daily lives have already been deeply permeated and improved by client-based inference techniques.

%% file: 5_client_training.tex
\section{Client-Based Training} \label{sec:client_training}
Client-based training is mainly motivated by the strong need of making the best use of user data generated on mobile devices and protecting users' privacy at the same time. Since we can hardly find any work and research paper about client-based local training, for the rest of this survey paper, client-based distributed training is referred to as client-based training for convenience. In contrast to traditional machine learning that require centralized datasets, client-based training uses mobile devices to solve machine learning problems according to local user data. The server aggregates all intermediate results and gets the trained model. Generally speaking, client-based training aims to transfer some computation tasks from centralized servers to decentralized mobile devices. It can not only offload servers' burden but also make use of the growing processing power on mobile devices. Furthermore, considering that data is processed locally in client-based training, the user data that is either too much to be uploaded or too sensitive to be uploaded can now participate in machine learning tasks. This gives possibility to improve the model's accuracy and accelerate the training process. More importantly, besides the benefits mentioned above, client-based training can even better preserve users' privacy. In this section, we first introduce and explain the motivations of client-based training in Section~\ref{subsec:client_training_motivations}. We formally define the task in Section~\ref{subsec:client_training_task_definiton}. Then we discuss the general constraints in Section~\ref{subsec:client_training_constraints}. The main challenges faced by client-based training will be claimed in Section~\ref{subsec:client_training_challenges}. Finally, we use federated learning and split learning as examples to show the current development of client-based training in Section~\ref{subsec:client_training_current_advances_federated_learning} and Section~\ref{subsec:client_training_current_advances_split_learning}.

\subsection{Motivations} \label{subsec:client_training_motivations}
We conclude motivations for client-based training as follows: (1) Client-based training keeps all advantages of client-based inference as client-based inference can be viewed as part of it; (2) In client-based training, the on-device model has a much shorter update cycle compared with server-based training and thus may perform better; (3) Client-based training is able to preserve user privacy and make full use of on-device sensitive user data at the same time; and (4) Some work in this direction has shown the feasibility and effectiveness of client-based training.

One problem faced by server-based training is that the cost~(communication cost and computation cost) is high. In server-based training, user data is generated on users' devices and then collected. However, in many cases, the data is so fine-grained and much that it incurs huge communication overhead when being transferred to the server. In addition, processing huge amounts of data and running large-scale machine learning on the server is not only time-consuming but also costly. Even if the model is finally trained to good performance on the server, the long training cycle will result in the delay of the model, which can cause decrease in its performance as users' behavior patterns may have already changed. Another problem in server-based training is the privacy risk. Specifically, the server in both centralized and distributed machine learning frameworks requires direct access to training data and thus need to store raw user data, which inevitably suffers outsider and insider attacks~\cite{popa2011cryptdb, popa2012cryptdb, popa2014building}. For example, a malicious hacker may invade the datacenter, compromise part of the server, and leak private databases. Further, if the server is untrusted, it may share user data with unauthorized entities or even trade for profits. To address these problems and disadvantages of server-based training, attempts on client-based training has been made.

\begin{figure}[t]
	\centering
	\subfloat[Client-Based Inference]{
	\begin{minipage}{.4\linewidth}
	\centering
	\includegraphics[width=\textwidth]{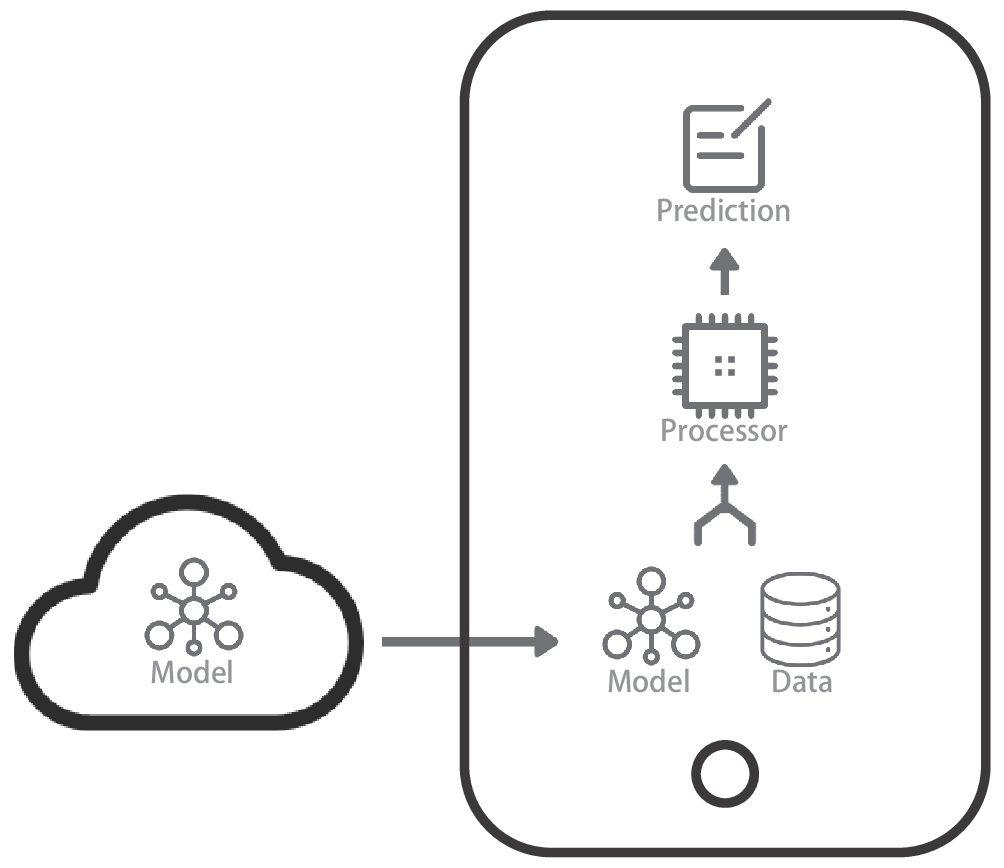}
	\end{minipage}
	}
	\begin{minipage}{.1\linewidth}
	\hspace*{\textwidth}
	\end{minipage}
	\subfloat[Client-Based Training]{
	\begin{minipage}{.4\linewidth}
	\centering
	\includegraphics[width=\textwidth]{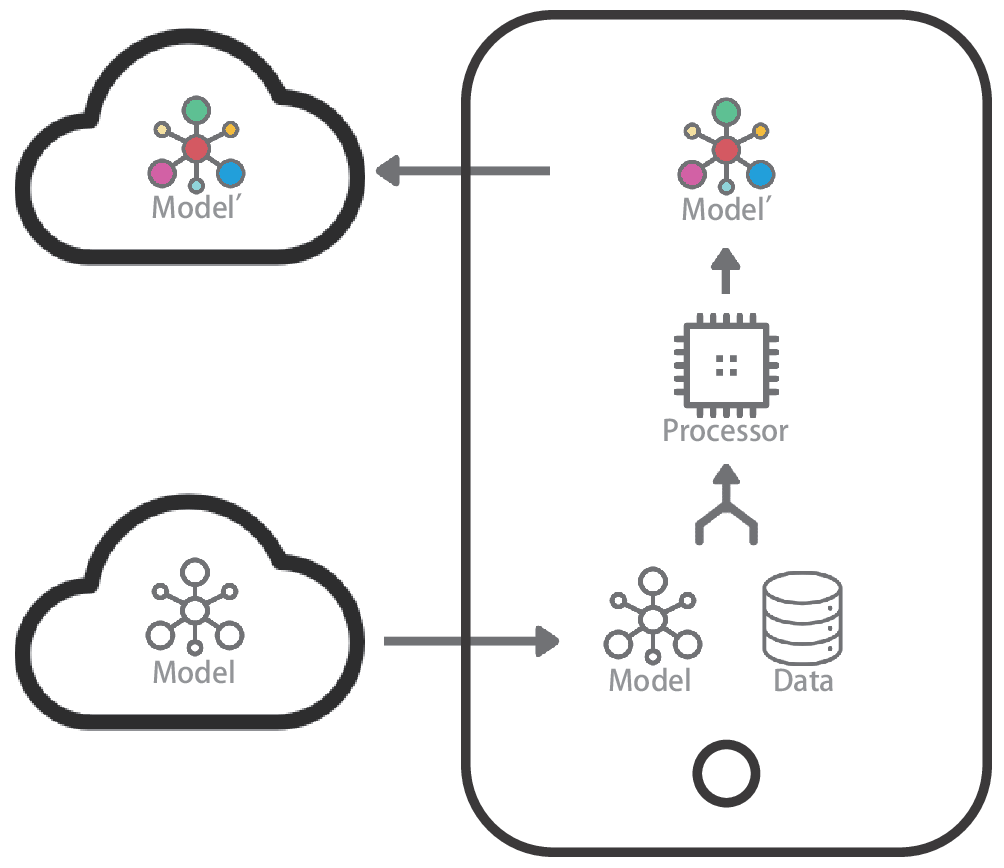}
	\end{minipage}
	}
	\caption{Illustrations of inference and distributed training in client-based machine learning.}
	\label{fig:inferenceandtraining}
\end{figure}

The difference between client-based inference and client-based training is shown in Fig.~\ref{fig:inferenceandtraining}. From the figure we can see that the client-based inference process just returns a prediction result which may be used to produce a service for the user locally. It has nothing to do for the server. However, if some part of the training process can be additionally done on-device, the generated trained model can not only become personalized but also provide valuable information for the global model stored on the server if the update of the local model is uploaded. This is exactly the idea of client-based training. As an example of client-based training, federated learning has already shown its effectiveness, which will be introduced later in Section \ref{subsec:client_training_current_advances_federated_learning}. Besides federated learning, some other works that are closely related to client-based training have also been proposed. He~\et~\cite{DBLP:journals/corr/abs-1808-04883} designed a framework for decentralized on-device linear learning. Koloskova~\et~\cite{DBLP:journals/corr/abs-1902-00340, DBLP:journals/corr/abs-1907-09356} further studied decentralized deep learning with compressed communication. They together show us the feasibility of on-device client-based training.

\subsection{Task Definition} \label{subsec:client_training_task_definiton}
We first introduce some notations and definitions in client-based training: (1) A real matrix $\mathbf{W} \in \mathbb{R}^{d_1 \times d_2}$ is the model learned from decentralized data; (2) A list $\{C_1, C_2, \dots, C_m\}$ contains all $m$ clients that are run on different mobile devices. They are also known as the workers in client-based training; (3) A list $\{D_1, D_2, \dots, D_m\}$ contains all local datasets used by the clients; and (4) A server $S$ that communicates with clients and arranges their tasks. $S$ can either be a single computer or a server cluster whose structure is transparent to the clients.

\begin{figure}[t]
  \centering
  \includegraphics[width=0.7\columnwidth]{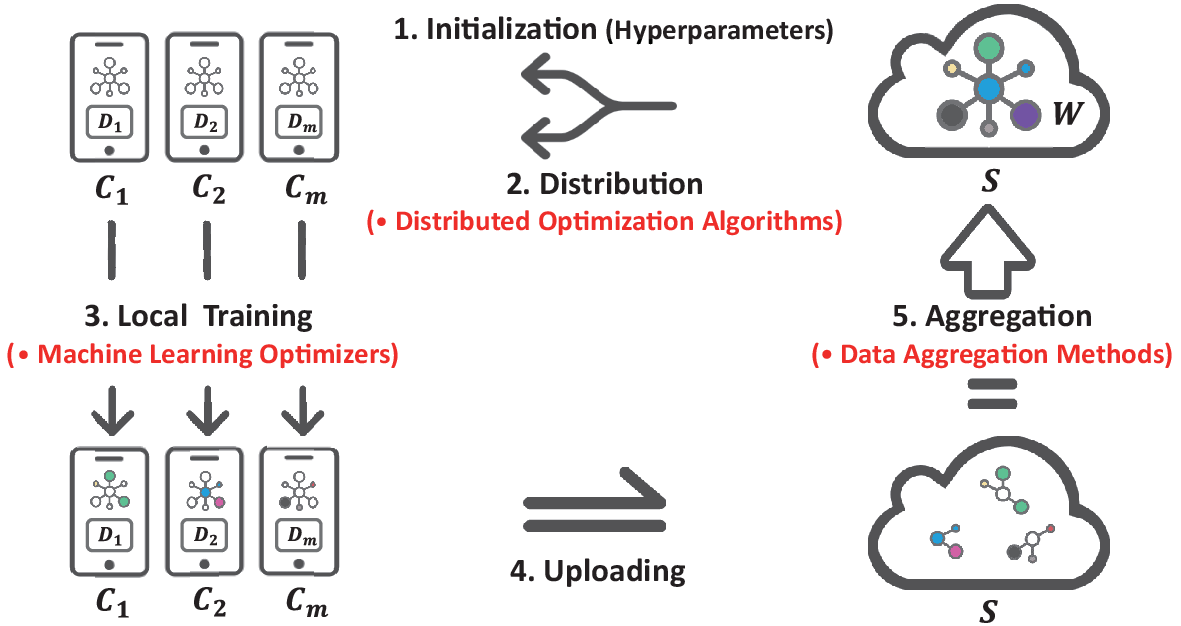}
  \caption{Workflow of client-based training.}
  \label{fig:general_process}
\end{figure}

\subsubsection{Workflow}
As shown in Fig.~\ref{fig:general_process},  a general workflow of client-based training consists of five phases: (1)~{\em Initialization:}~The workers $\{C_1, C_2, \dots, C_m\}$ run on mobile devices download hyperparameters from the server $S$ and get themselves initialized; (2)~{\em Distribution:}~The server $S$ arranges the tasks and sends them to the workers $\{C_1, C_2, \dots, C_m\}$. The model to be trained should also be distributed to the workers in this step; (3)~{\em Local Training:}~The workers $\{C_1, C_2, \dots, C_m\}$ train the local models based on the local datasets $\{D_1, D_2, \dots, D_m\}$. Updates will be generated after a certain number of local training iterations; (4)~{\em Uploading:}~The workers $\{C_1, C_2, \dots, C_m\}$ upload the updates to the server $S$; and (5)~{\em Aggregation:}~The server $S$ aggregates all updates to generate a final update and applies it to the global model $\mathbf{W}$. A single round of learning ends here. Go back to step 2 to start a new round of learning. A similar workflow has been adopted in federated learning protocol~\cite{bonawitz2019towards}.

\subsubsection{System Modules}
To accomplish client-based training, the system is made up of three modules: local training module, communication module, and aggregation module. We show how they cooperate with each other and what kinds of techniques can be used to improve them.

The local training module is implemented on the worker. It works in step 3. Besides the data preprocessing part, the core component of the local training module is the local machine learning optimizer. Considering that the resources available for client-based training are very limited, the local optimizer must be chosen and designed carefully. It should be run on mobile devices without significant bad effect on users' daily usage experience. For the local training module, the input is the initialized model and the output is the update. The update can be generated based on either the new model which has been trained on the local dataset or the difference between the two models.

The Communication module takes part in step 1, step 2 and step 4. It is implemented on both the server and the worker, which means the server and the worker have to cooperate with each other for communication. The communication module is responsible for two jobs: task arrangement and data transfer. For the server, it should concern about the task arrangement part whose main purpose is to divide the original complicated problem into easier sub-problems. This can be done through distributed optimization algorithms. The input is the original problem and the output is a set of sub-problems. From the worker's perspective, data transfer is the main challenge as mobile network resources are limited and expensive. Compression methods can be used to reduce the communication cost. Here, the input is the original data and the output is the compressed data.

The aggregation module is implemented on the server. It works in step 5. Since the number of updates received from workers can be very large, it is inefficient and even impossible to apply all updates to the global model one by one. Thus, the main job of the data aggregation module is to extract and make the best use of the information contained in the updates. It takes all received updates as inputs. Its output can be either a final update which can be directly applied to the global model or a new model which replaces the old one.

\subsection{Constraints} \label{subsec:client_training_constraints}
The only difference between server-based training and client-based training is where the training process is done. However, this change introduces additional constraints for the whole system due to the limitations on mobile devices: (1) Compared with PCs and servers, the hardware resources available on mobile devices are poor and limited; (2) The network condition for mobile devices is unstable, and the communication cost is very high; and (3) Unlike servers which are owned and managed by data centers, mobile devices are not totally under control, thus the reliability and the security of the whole decentralized training process are not guaranteed.

The first limitation is similar to the first challenge of client-based inference as both of them require plenty of computation. The second limitation is introduced considering that client-based training additionally needs the communication process to accomplish distributed training. The third limitation indicates that client-based training is more vulnerable to attacks. In what follows, we show how these three limitations lead to six concrete constraints of client-based training.

\paragraph{\textbf{Simple On-Device Task}} The machine learning sub-problems solved on mobile devices should be easy. This constraint is desired according to the first limitation. Although the development of mobile devices has kept accelerating in recent years, their computation power is still far less than that of personal computers, not to mention the server clusters. The key reason for this phenomenon is that mobile devices need to be portable. In this situation, chipsets are specially designed for mobile devices to balance between the need for lower energy consumption and the need for greater power. Thus, unlike the CPUs used on personal computers which can easily reach 100W in power with stable power supply, the peak power of CPUs used on mobile devices is usually under 10W. Considering that mobile devices have to spare some resources and energy to support its basic functions, the part which can be used for supporting mobile machine learning becomes even less. Moreover, the mobile devices usually have no additional cooler and just use passive cooling methods. This also limits the maximum power of them. Considering the facts given above, the necessity of the simple on-device task constraint has been clear. First, mobile devices cannot handle complex machine learning tasks because the power is limited. Second, since only passive cooling methods are used, a hard machine learning task is very likely to cause the mobile device becomes hot. This must not happen as it has huge negative effect on user experience. Therefore, the machine learning sub-problems solved on mobile devices should be easy enough so that the users are even unaware of its running at all.

\paragraph{\textbf{Short Training Time}} The time spent on one round of local on-device training should be short. We have mentioned that the hardware resources available on mobile devices are poor and limited compared with personal computers. Thus, the systems run on mobile devices are specially designed. Background apps may be paused by the system to save resources for the foreground app. Considering that most users just use an app for a little while at one time, it is better to finish the training task during the time that the app is running in the foreground. Otherwise, if the training has not been finished when the app is paused or quit, the generated update may not be sent to the server in time and become useless at the next start due to the long delay.
    
\paragraph{\textbf{No Uploading}} Local private user data should not be uploaded to the server in principle. For now, although the 4G network has already been deployed widely and the 5G network is coming, the cost of data on mobile devices is still expensive. What's more, local data on mobile devices may contain sensitive user information, which means it is not suitable for being uploaded to the server. Thus, to best preserve user privacy, the local raw user data should not be uploaded in principle. However, considering that uploading a small subset of local data is feasible and may be helpful for the aggregation process on the server, it can be permitted under special situations with guaranteed privacy. Note that {\em No Uploading} is one important cause of the significant difference between server-based distributed training and client-based training. Without uploading data, the local datasets used for training are specific to their owners and can never become Independent~and~Identically~Distributed~(IID) which is a necessary condition for many machine learning algorithms. Moreover, {\em No Uploading} also causes the sizes of local datasets to be unbalanced, and results in the information contained in different workers' updates being unbalanced. Hence, how to determine which updates are more valuable~(contains more information) becomes another problem to be solved.

\paragraph{\textbf{Low Communication Overhead}} The data transferred in one round of learning should not be too much. This constraint also aims at protecting the experience of users' daily usage. If too much data has to be transferred, it may be a heavy burden for the mobile devices even if they are connected to WLANs and can cause other apps that are using the network to be blocked.
    
\paragraph{\textbf{Low Communication Frequency}} The communication frequency should be low and it is better to let the workers decide when to communicate. As the network condition for mobile devices is unstable, a scheme with frequent communication is unsuitable. What's more, a low communication frequency with a long training time is beneficial for reducing the communication cost. The reason why we suggest letting the workers decide when to make the communication is that the high heterogeneity of workers makes the server have no idea when the sub-problems will be solved. By letting workers make the decision, we may also reduce the communication cost as workers can choose to do the communication at the time when they are connected to WLAN.

\paragraph{\textbf{High Awareness of Data Reliability}} The server should be highly aware of the data reliability of the received updates. Since the decentralized training scheme cannot guarantee the reliability of the received data, client-based training suffers from fake updates or low-quality updates. To prevent machine learning poisoning attacks, a possible solution is to let the server evaluate the data reliability by making use of update diversity or additional information such as worker reputation. With high awareness of data reliability, the server can distinguish and reject low-quality updates and make the training process become more secure and stable.

\subsection{Challenges} \label{subsec:client_training_challenges}
From the above six constraints, we can summarize the major challenges of client-based training: (1) Transforming original machine learning tasks to easier sub-problems which can be solved on mobile devices effectively and efficiently; (2) Dealing with the unbalanced non-IID dataset and cover its negative effects; (3) Compressing the transferred data between the server and the clients; (4) Lowering communication frequency and overhead; and (5) Ensuring the efficiency and security of data reliability evaluation.

For challenge (1), although we can refer to distributed optimization algorithms, none of them guarantees that it performs well with millions or even billions of workers. The huge number of workers and the limited resources available on workers may cause the optimization problem to become so difficult that we must design a brand new solution for it.

For challenge (2), some existing methods (\eg, \cite{DBLP:journals/corr/abs-1806-00582}) use data-sharing strategies to mitigate the negative effect of the unbalanced non-IID dataset. The problem of data-sharing strategies is that raw data transfer can be costly and risky for mobile devices. Another possible solution is to design new machine learning algorithms which are insensitive to the data distribution. For example, semi-cyclic SGD~\cite{10.5555/3327757.3327856} can adapt to data with block-cyclic pattern.

Regarding challenge (3) and (4), they have already been widely studied in other areas (\eg, data compression and computer network). However, considering that general methods normally may not well adapt to every scenario, specific algorithms that are closely combined with client-based training must be more effective and thus are needed urgently.

For challenge (5), to prevent model poisoning attacks~\cite{DBLP:journals/corr/abs-1807-00459, shayan2018biscotti, fung2018mitigating}, some existing studies~\cite{fung2018mitigating} use update diversity to distinguish low-reliability updates. Some others~\cite{kang2019incentive, kang2020reliable} choose to calculate the reputation of workers and use the reputation score to represent the data reliability. Techniques such as blockchain have been used to achieve secure reputation management. Although these methods can effectively protect the global model training process from the attacks, they incur large computational cost as the number of workers increases. Thus, the main task here is to design an algorithm which can guarantee both security and efficiency.

To better learn about these challenges and their possible solutions, we use federated learning and split learning as two examples and introduce their current advances.


\subsection{Current Advances in Federated Learning} \label{subsec:client_training_current_advances_federated_learning}
In recent years, attempts at the implementation of client-based training have occurred. Federated Learning, as an example, was proposed by researchers from Google in 2015. Generally speaking, federated learning means to do machine learning tasks federatedly among a large number of mobile devices with on-device private data protected. Its core idea is to train the model locally on user devices and aggregate updates on the server without uploading raw user data, which shares a similar scheme of distributed machine learning. From 2016 to 2018, Google published several related articles to complement federated learning's framework. In 2019, applications of federated learning appear. A detailed survey~\cite{2019arXiv191204977K} about federated learning is given in December, 2019. Federated learning can be viewed as an attempt at client-based training with privacy preservation as the primary goal. Existing federated learning work mainly focuses on studying how to improve the final model's performance and how to design an efficient communication scheme. In this section, problems and current advances in federated learning are introduced and discussed.

\subsubsection{Model Performance}

\paragraph{\textbf{Literature Review}}
The idea of federated learning can be traced back to Distributed~Selective~Stochastic~Gradient~Descent~(DSSGD)~\cite{shokri2015privacy}. It was published in October 2015 and is about distributed deep learning without sharing datasets.

Following the idea of distributed learning and privacy preservation, in November 2015, researchers from Google submitted their first attempt on federated learning~\cite{konevcny2015federated}. Federated learning can be viewed as an improved version of DSSGD which is optimized for mobile devices. In this work, three basic properties of federated learning's scenario were given: (1)~\textit{Non~Independent~and~Identically~Distributed~(non-IID) Data}; (2)~\textit{Unbalanced Data}; and (3)~\textit{Massively Distributed Data}. (Note that the fourth property (4)~\textit{Limited Communication} was introduced in~\cite{mcmahan2016communication} and will be discussed in Section~\ref{subsubsec:communication_efficiency}.) Also, this work proposed an efficient federated optimization algorithm called Distributed~Stochastic~Variance~Reduced~Gradient~(DSVRG) based on SVRG~\cite{johnson2013accelerating} and DANE~\cite{shamir2014communication}.

Later in 2016, Kone{\v{c}}n{\`y}~\et~\cite{konevcny2016federated} complemented and improved the above work. DSVRG was renamed as Federated~SVRG~(FSVRG). Equations and mathematical proofs for it were also provided.

After the above works had formulated the basic training scheme of federated learning, researchers continued to study the effectiveness of federated learning and tried to improve its performance. Zhao~\et~\cite{DBLP:journals/corr/abs-1806-00582} analyzed the negative effect of non-IID datasets on model performance and provided a simple data sharing strategy to deal with it. Yu~\et~\cite{yu2019parallel} studied on why model averaging works for deep learning tasks. Eichner~\et~\cite{pmlr-v97-eichner19a} discovered that cyclic patterns in the data samples is hard to be avoided in federated learning and does harm to the performance of SGD. They proposed Semi-Cyclic SGD to correct this problem when optimizing convex objectives. Mohri~\et~\cite{pmlr-v97-mohri19a} proposed a new framework of agnostic federated learning to avoid the federated model being biased towards different clients. Chen~\et~proposed FedMeta~\cite{DBLP:journals/corr/abs-1802-07876} which combines federated learning with meta-learning. To better evaluate various kinds of federated learning algorithms, LEAF~\cite{DBLP:journals/corr/abs-1812-01097}, a modular benchmarking framework for learning in federated settings, was proposed. It contains open-source federated datasets and is continuously updated.

\paragraph{\textbf{Discussion}}
While federated leaning training algorithms have developed from DSSGD to FSVRG, the most important non-IID data problem has not been solved well yet. What's more, the cyclic pattern of data is also harmful and should be taken into consideration. In conclusion, to further improve the model performance of federated learning, we need to deal with the {\em cyclic unbalanced non-IID data} problem under the {\em non-convex objective} condition.

\subsubsection{Communication Efficiency} \label{subsubsec:communication_efficiency}

\paragraph{\textbf{Literature Review}}
The first aspect of improving communication efficiency is reducing communication rounds. In~\cite{mcmahan2016communication}, Federated Averaging~(FedAvg) was proposed to deal with the fourth basic characteristics of federated learning \textit{Limited Communication}. By enabling multiple local training iterations and using model averaging methods, the number of communication rounds is reduced. Similarly, based on parallel restarted SGD~\cite{yu2019parallel}, Yu~\et~\cite{pmlr-v97-yu19d} proposed parallel restarted SGD with momentum to enlarge local training steps and thus reduce the total number of rounds.

Meanwhile, reducing the size of communication data also helps improve communication efficiency. According to~\cite{konevcny2016federated2}, as on most occasions the bandwidth of the uplink is much poorer than that of the downlink, reducing the uplink communication cost is a more urgent task. Two kinds of approaches~({\it Structured Updates} and {\it Sketched Updates}) which can lower the size of the updates are implemented and tested in federated learning in~\cite{konevcny2016federated2}. Besides the general compression methods, Caldas~\et~\cite{DBLP:journals/corr/abs-1812-07210} proposed Federated Dropout. With Federated Dropout, each worker trains a smaller sub-model instead of the whole global model. Then the size of updates is also reduced.

\paragraph{\textbf{Discussion}}
Although the above-mentioned methods do improve the communication efficiency, they are still using the two-tier server-worker communication structure. This scheme not only brings huge communication burden to the central server, but also suffers from the instability of mobile workers' network. Perhaps we can consider using a multi-tier communication structure to further improve both communication efficiency and communication stability.

\subsubsection{Security \& Privacy}

\paragraph{\textbf{Literature Review}}
After the development of FedAvg~\cite{mcmahan2016communication}, to better ensure the security of the aggregation process in federated learning, Google proposed a practical secure aggregation method~\cite{10.1145/3133956.3133982} for privacy-preserving machine learning. In the secure aggregation process, the secure multiparty computation technique is used to compute sums of model parameter updates. What's more, model poisoning attacks~\cite{DBLP:journals/corr/abs-1807-00459, shayan2018biscotti, fung2018mitigating} toward federated learning have also been studied. The main purpose of model poisoning is to make the trained model output wrong answers or even attacker-chosen answers. To prevent model poisoning attacks, update diversity~\cite{fung2018mitigating} and worker reputation~\cite{kang2019incentive, kang2020reliable} can be used to recognize unreliable updates.

For better privacy preservation, McMahan~\et~\cite{DBLP:journals/corr/abs-1710-06963} applied differential privacy methods to FedAvg and only resulted in a negligible cost in inference accuracy. Agarwal~\et~\cite{10.5555/3327757.3327856} proposed cpSGD to achieve both differential privacy and communication efficiency in federated learning settings. Niu~\et~\cite{10.1145/3372224.3419188} proposed a secure federated submodel learning
scheme with tunable property which enables the workers to tune privacy and utility. Considering that the maximum contribution is an important parameter for differential privacy algorithms~(the noise to be added to data is closely related to it), Amin~\et~\cite{pmlr-v97-amin19a} studied how to bound user contributions in federated learning. Yang~\et~\cite{DBLP:journals/corr/abs-1901-08755} proposed a novel lossless privacy-preserving tree-boosting system called SecureBoost.

\paragraph{\textbf{Discussion}}
Regarding better privacy, although federated learning can naturally protect sensitive raw user data, we still need to make sure that the attacker cannot infer information from the transferred updates. For the security part, since federated learning distributes the training process to unreliable mobile devices, attacks which aim at misleading the final model can be more easily implemented. For example, we should pay more attention to the potential data poisoning attacks.

\subsubsection{Applications} Federated learning has already been used in some applications and shown satisfying performance. Google first applied federated learning to Gboard to improve its query suggestions~\cite{DBLP:journals/corr/abs-1812-02903}. They tested federated learning on 100 clients. Results shew that federated learning do improve the performance of the deployed LSTM model. After that, they also applied federated learning to the mobile keyboard next-word prediction task~\cite{DBLP:journals/corr/abs-1811-03604}. According to the evaluation done on server-hosted logs data, the federated-trained CIFG model performs nearly as well as the centralized-trained CIFG model. And for evaluation done on client-owned data caches, federated learning even outperforms centralized learning. Google has also applied federated learning to learn Out-Of-Vocabulary~(OOV) words~\cite{DBLP:journals/corr/abs-1903-10635}. This work conducted both simulated federated learning on a non-IID dataset and real-word federated learning on data hosted on user mobile devices. Results shew that the federated learning method can learn OOV words effectively. Google introduced their system design for federated learning at scale in~\cite{bonawitz2019towards}, which mainly focus on how to design an elastic parameter server to support large number of clients in real-word settings.

Besides Google, many other researchers were also exploring federated learning. Smith~\et~\cite{NIPS2017_7029} combined federated learning with multi-task learning. Intel~\cite{sheller2018multi} used federated learning to do multi-institutional deep learning for brain tumor segmentation without sharing patient data. Yang~\et~\cite{DBLP:journals/corr/abs-1812-03337} proposed Federated Transfer Learning~(FTL). This work shew that federated learning is also suitable for being applied to machine learning tasks in the scenario of cooperation among banks where sensitive information mustn't be shared. Felix~\et~\cite{DBLP:journals/corr/abs-1903-02891} proposed Sparse Ternary Compression~(STC). STC is more robust to non-IID datasets and works as a substitute for FedAvg.

\subsubsection{Surveys} Google has published a very detailed survey paper~\cite{2019arXiv191204977K} to summarize all works and researches related to federated learning. Advances of federated learning and open problems in this area have also been discussed. Yang~\et~provided a survey~\cite{yang2019federated} about federated learning's concept and applications. In this survey paper, Federated Learning is extended and classified into Horizontal~Federated~Learning~(HFL) and Vertical~Federated~Learning~(VFL). Li~\et~\cite{li2019federated} write a survey to discuss challenges, methods, and future directions of federated learning.

\subsection{Current Advances on Split Learning} \label{subsec:client_training_current_advances_split_learning}
Just like federated learning can be viewed as a special case of data parallel distributed learning where datasets are distributed on mobile workers, split learning can be considered as a special case of model parallel distributed learning where the model is distributed on the server and mobile workers. Here, we continue to introduce current advances in split learning.

\subsubsection{Model Performance} \label{subsubsec:split_model_performance}

\paragraph{\textbf{Literature Review}}
The idea of split learning first appeared in~\cite{DBLP:journals/corr/abs-1810-06060} and was proposed by a research team from MIT. It is motivated by the need of using multiple agents to collaboratively train a deep neural network without transferring raw sensitive user data, which is very similar to the motivation of federated learning. To achieve this goal, Split Neural Network~(SplitNN) was designed. In SplitNN, only bottom parts of the model are trained on an worker who owns the raw data. The gradients together with the labels for the training data are transmitted to the server who accomplishes the rest training process for the top parts of the model. Considering that the labels may also reveal sensitive user information, U-shaped SplitNN is designed. In U-shaped SplitNN, the server only processes the middle layers of the model. The bottoms layers and the top layers are stored on workers to process the raw data and the corresponding labels. Experiments have been done to demonstrate the effectiveness of SplitNN.

Later, the concept of split learning was formally proposed in~\cite{DBLP:journals/corr/abs-1812-00564}. SplitNN, as a method of split learning, has been improved in this work. Several possible privacy-preserving structures of SplitNN have been provided and discussed. We will introduce this part later in Section \ref{subsubsec:split_privacy}. By comparing SplitNN with large-scale SGD and FedAvg on CIFAR 10 and CIFAR 100, it is shown that SplitNN can achieve higher validation accuracy with much less computation. 

\paragraph{\textbf{Discussion}}
In SplitNN, although a training algorithm using multiple workers have been designed, the whole training process is still run sequentially. According to the algorithm, a worker must fetch the latest model before it starts training. After a training iteration, this worker's updated model will be marked as the latest model. This scheme implies that only one worker can train the latest model at the same time. Although Singh~\et~\cite{singh2019detailed} proposed an approach called ``split learning without any client weight synchronization'', how it works and how it performs are not described. On the other hand, since there is no model aggregation process, multiple workers training the latest model in parallel will result in conflicts. This lack of parallel training scheme makes SplitNN fail to get benefit from parallel training acceleration. Moreover, while local raw data is not transferred in split learning, the impact of local non-IID dataset has not been analyzed yet.

\subsubsection{Communication Efficiency}
\paragraph{\textbf{Literature Review}}
A detailed comparison of communication efficiency of split learning and federated learning was given in~\cite{singh2019detailed}. The theoretical analysis shows that split learning becomes more communication-efficient as the number of clients increases and can well adapt to big models. One shortcoming of this work is that the theoretical analysis has not been validated by experiments.

\paragraph{\textbf{Discussion}}
The communication efficiency of split learning has not been well studied yet. As workers need to transfer data with the server in each training iteration, the network condition~(\eg, bandwidth and latency) can greatly influence the system efficiency. What's more, the applicability of communication compression methods needs to be further examined.

\subsubsection{Security \& Privacy} \label{subsubsec:split_privacy}
\paragraph{\textbf{Literature Review}}
Attacks on split learning have been studied in~\cite{abuadbba2020can}. This work shows that it is possible to reconstruct the raw data from the worker's outputs if only a few convolution layers is trained on the workers. In addition, neither introducing additional hidden layers nor applying differential privacy to the split layer can effectively mitigate this shortcoming. Both of the two attempts cause an unacceptable decrease in the model accuracy.

Several privacy-preserving structures of SplitNN have been introduced in~\cite{DBLP:journals/corr/abs-1812-00564}. As multiple institutions might own different modalities of the same user's data, vertical split learning is designed to deal with this kind of data which is partitioned by features. Extended vanilla split learning and Tor-like multi-hop split learning arrange additional workers to process the middle layers of the model. This structure helps to cover the identities of the bottom workers who use their sensitive raw data to train the bottom layers, and thus the privacy of the bottom workers are better preserved by the anonymity. Besides the basic structure design, Vepakomma~\et~\cite{vepakomma2019reducing} used two losses in one model to reduce data leakage for SplitNN. Sharma~\et~~\cite{sharma2019expertmatcher2} proposed ExpertMatcher to do model matching for split learning with only the encoded hidden representation of local raw data shared. By hiding the raw data representation, user privacy is better preserved.

\paragraph{\textbf{Discussion}}
One important problem not studied yet is how to trade off between on-device model complexity and user privacy. By putting some complex layers of the model on workers, the privacy may be better preserved as it becomes harder to infer raw data. Meanwhile, split learning also suffers model poisoning attacks. Now that some parts of the model even do not exist on the server, model poisoning attacks become much easier and should be more carefully handled.

\subsubsection{Applications}
Split learning has been tested in the medical field in~\cite{poirot2019split}. U-shaped SplitNN has been implemented to enable the collaborative machine learning between several hospitals. Experiments are done on two medical datasets: retinal fundus photos and chest X-rays. Results show that split learning outperforms non-collaborative methods greatly.

\subsubsection{Surveys}
Vepakomma~\et~\cite{DBLP:journals/corr/abs-1812-03288} have surveyed methods for deep learning without revealing raw data, including large batch SGD, federated learning, and split learning. Key ideas, limitation, and future trends of them have been simply discussed. In addition, this survey paper has introduced several cryptographic techniques which can be used to further preserve privacy in machine learning area, including homomorphic encryption, oblivious transfer, and garbled circuits.

%% file: 6_future_directions.tex
\section{Future Directions} \label{sec:future_directions}

In this section, we introduce some potential research directions of client-based training. Most of them are motivated by the problems and challenges discussed in Section \ref{sec:client_training} that have not been studied well (\eg, Non-IID Training Sets). Others are ideas which can improve the robustness or the adaptivity of the system (\eg, General Mobile Training Framework).

\paragraph{\textbf{Non-IID Training Sets}} This problem was first introduced in federated learning. It also exists in the scenario of other client-based training methods such as split learning as we can no longer upload the original data and do a shuffle. Zhao~\et~\cite{DBLP:journals/corr/abs-1806-00582} has shown the negative effect of non-IID datasets on model convergence and proposed a data sharing strategy to deal with it. However, it may be difficult for mobile devices to share a small part of the local dataset with others because the process is hard to be managed and the communication cost can be high. We also have to ensure that the privacy is still preserved during the whole data transmission process. Another possible solution is to develop new machine learning algorithms which are not sensitive to the distribution of training set. However, this direction does not have much existing work that can be referenced and may become a new hard machine learning problem.
    
\paragraph{\textbf{Aggregation Methods}} For existing data aggregation methods designed for server-based distributed training (\eg, FedAvg~\cite{mcmahan2016communication}, Ensemble-Compression~\cite{sun2017ensemble}, and Codistillation~\cite{anil2018large}), none of them guarantees a fast convergence. Since the number of workers is usually under one hundred in server-based distributed training, no one has the experience to aggregate tens of thousands of updates in a round in client-based training. Moreover, the above-introduced methods all have disadvantages. The averaging-based methods may cause a significant decrease in the convergence speed as they reduce the scale of updates on weights. They may also not be suitable for non-convex problems. The distillation-based methods may not be able to merge tens of thousands of models because they need much additional computation to handle this complex task. Thus, how to design an effective, efficient, and robust aggregation method is an urgent problem to be solved.

\paragraph{\textbf{Security \& Privacy}.} Although client-based training can prevent the leakage of raw user data, Shokri~\et~\cite{shokri2017membership} have already shown the feasibility of the attack against machine learning models. This work demonstrated how to inference user membership only based on the trained model. Differential privacy, as a solution to this problem, requires additional computation on mobile clients and can cause decrease in model accuracy. Secure aggregation~\cite{10.1145/3133956.3133982} does no harm to the model performance but can hardly be deployed in large-scale client-based training due to its high complexity. On the other hand, malicious clients may adjust their update data in order to affect or control the behavior of the final model and gain benefits for themselves~\cite{DBLP:journals/corr/abs-1807-00459}. These facts imply that better privacy and security in the scenario of client-based training is still an open problem.

\paragraph{\textbf{Communication with 5G}} The 5th generation mobile network~(5G) brings high-bandwidth and low-latency network to mobile devices. The training slow-down caused by network latency and the client drop-off caused by network instability can be greatly relieved by 5G. Since communication efficiency is a key concern for the implementation of client-based training, the high bandwidth, low latency, and good stability of 5G may help client-based training become more robust~\cite{loghin2020disruptions}.

While current client-based training methods are commonly using server-client communication scheme~(\eg, parameter server), 5G provides us with the opportunity to extend the communication scheme to device-to-device~(D2D) and vehicle-to-vehicle~(V2V). Some work~(\eg, \cite{samarakoon2018federated}) has already tried to combine federated learning with V2V networks to achieve low-latency neighbor cooperation. We can also use 5G D2D network~\cite{savazzi2020federated} to add middle layers between the server and the clients to realize multi-tier network structure which is more robust and scalable.

What's more, as 5G has brought stronger connectivity to mobile devices, secure data utilization strategy becomes an urgent problem. Thus, besides being improved by 5G techniques, client-based training has also been used to deal with 5G problems such as Network Data Analytics Function~(NWDAF)~\cite{niknam2019federated, isaksson2020secure}. These facts show that combining client-based learning with 5G is beneficial to both sides and thus will be an important future direction.

\begin{table*}[!t]
  \renewcommand\arraystretch{1.3}
  \centering
  \caption{Available datasets for client-based training in LEAF.}
  \label{tab:datasets_leaf}
  \resizebox{0.8\textwidth}{!}{
    \begin{tabular}[t]{|c|c|c|c|c|c|}
    \hline
    \multirow{2}{*}{\textbf{Name}} & \multirow{2}{*}{\textbf{\#samples}} & \multirow{2}{*}{\textbf{\#users}} & \multicolumn{2}{|c|}{\textbf{\#samples per user}} & \multirow{2}{*}{\textbf{Task}}\\
    \cline{4-5}
    ~ & ~ & ~ & \textbf{mean} & \textbf{stdev} & ~\\
    \hline
    \textbf{FEMNIST} & 805,263 & 3,550 & 226.83 & 88.94 & Image Classification\\
    \hline
    \textbf{Shakespeare} & 4,226,158 & 1,129 & 3,743.28 & 6,212.26 & Sentiment Analysis\\
    \hline
    \textbf{Twitter} & 1,600,498 & 660,120 & 2.42 & 4.71 & Next-Character Prediction\\
    \hline
    \textbf{CelebA} & 200,288 & 9,343 & 21.44 & 7.63 & Image Classification\\
    \hline
    \textbf{Reddit} & 56,587,343 & 1,660,820 & 34.07 & 62.9 & Next-word Prediction\\
    \hline
    \textbf{Synthetic Dataset} & 107,553 & 1,000 & 107.55 & 213.22 & Classification\\
    \hline
    \end{tabular}%
    }
\end{table*}%

\paragraph{\textbf{Standardization \& Benchmark}} Up till now, there still does not exist a white paper to comprehensively define and claim the standard of client-based training. Since deploying client-based training requires a trade-off between a lot of properties (\eg, model accuracy, communication overhead, complexity, privacy, and the scale of supported clients), the comparison between different client-based training algorithms will be difficult without pre-defined standards and evaluation metrics. Moreover, as client-based training can be applied to various kinds of tasks, it is also important to collect and release corresponding benchmark datasets which satisfy the client-based training settings. We recall that the {\em No Uploading} constraint in client-based training forces the local datasets to be unbalanced non-IID datasets which are partitioned by users. LEAF~\cite{DBLP:journals/corr/abs-1812-01097} can be taken as a reference. Table~\ref{tab:datasets_leaf} has listed the currently available datasets in LEAF. All these datasets allow partition by user, which means they can be used to test client-based training algorithms under the non-IID condition. For now, LEAF datasets have covered only a few tasks. More benchmark datasets are needed to support the development of client-based training in different areas.

\paragraph{\textbf{Deployment Scenarios}} Since both federated learning and split learning are general client-based training frameworks which concern about the overall learning process and scheme, they have to be combined with concrete machine learning methods and models when being applied to real-life applications. However, as they are newly emerging techniques and haven't been tested on many tasks, the suitable deployment scenarios for any of them are still not clear. For now, client-based training has caught people's eyes because it can guarantee user data privacy sacrificing a little model performance and training speed. That is the reason why federated learning and split learning have been applied to applications whose user data is sensitive (\eg, Gboard) and the health area. Considering that client-based training is experiencing rapid development with its performance, efficiency, and robustness all being improved, it is urgent to figure out what else can client-based training do and how to better deploy it on mobile devices in various kinds of real-life scenarios.
    
\begin{table*}[!t]
  \renewcommand\arraystretch{1.3}
  \centering
  \caption{Mobile machine learning frameworks.}
  \label{tab:mobile_frameworks}
  \resizebox{0.67\textwidth}{!}{
    \begin{tabular}[t]{|c|r|r|r|r|c|}
    \hline
    \textbf{Name} & \tabincell{r}{\textbf{(Server-Based)}\\ \textbf{Centralized}} & \tabincell{r}{\textbf{(Server-Based)}\\ \textbf{Distributed}} & \tabincell{r}{\textbf{Mobile}\\ \textbf{Inference}} & \tabincell{r}{\textbf{Mobile}\\ \textbf{Training}} \\
    \hline
    \textbf{TensorFlow Lite}~\cite{TensorFlowLite} & TensorFlow \Checkmark & TensorFlow \Checkmark & \Checkmark & \XSolidBrush\\
    \hline
    \textbf{Core ML}~\cite{CoreML} & Create ML \Checkmark & \XSolidBrush & \Checkmark & \XSolidBrush \\
    \hline
    \textbf{PyTorch Mobile}~\cite{PyTorch} & PyTorch \Checkmark & PyTorch \Checkmark & \Checkmark & \XSolidBrush \\
    \hline
    \textbf{NCNN}~\cite{NCNN} & \XSolidBrush & \XSolidBrush & \Checkmark & \XSolidBrush \\
    \hline
    \textbf{Paddle Lite}~\cite{PaddleLite} & Paddle \Checkmark & Paddle \Checkmark & \Checkmark & \XSolidBrush \\
    \hline
    \textbf{MNN}~\cite{alibaba2020mnn} & MNN \Checkmark & \XSolidBrush & \Checkmark & \Checkmark \\
    \hline
    \end{tabular}%
    }
\end{table*}%

\paragraph{\textbf{General Mobile Training Framework}} We list some commonly used mobile machine learning frameworks in Table~\ref{tab:mobile_frameworks}. Although the computation power of mobile devices is already sufficient for training small models, many existing open-source mobile machine learning frameworks such as TensorFlow~Lite~\cite{TensorFlowLite} and PyTorch~Mobile~\cite{PyTorch} still only support inference operations, which means they are actually mobile {\em inference} frameworks. Without a general mobile training framework, implementing client-based training on mobile applications can be inefficient and time-consuming because developers have to realize all training operations by themselves for each task. To deal with this problem, MNN~\cite{alibaba2020mnn} from Alibaba has provided an on-device training module. MNN supports constructing a model from zero and training it totally on mobile devices. We hope that other mobile machine learning frameworks can also add support for mobile on-device training.

%% file: 7_conclusion.tex
\section{Conclusion} \label{sec:conclusion}
In this survey, we have provided a thorough overview of the development of machine learning in recent years, from traditional server-based machine learning to emerging client-based learning. We have discussed their purposes and demonstrated the sufficiency and necessity of client-based machine learning. Specifically, for client-based inference, we have discussed its challenges and demonstrated its current advances, especially in the fields of computer vision and natural language processing. In addition, for client-based training, we have illustrated motivations and bottlenecks, given a clear task definition, and further offered a general guideline for practicers. As typical examples of client-based training, we have introduced the concepts of federated learning and split learning and also reviewed their current advances. We finally have pointed out some future research directions of client-based machine learning in both academia and industry. In summary, applying client-based machine learning to real-world industrial applications is still faced with many challenges and opportunities, which calls for more attention to be paid on its future development. We hope that this survey can be a good starting point.